\documentclass{article}

\usepackage{arxiv}

\usepackage[utf8]{inputenc} % allow utf-8 input
\usepackage[T1]{fontenc}    % use 8-bit T1 fonts
\usepackage{hyperref}       % hyperlinks
\usepackage{url}            % simple URL typesetting
\usepackage{booktabs}       % professional-quality tables
\usepackage{amsfonts}       % blackboard math symbols
\usepackage{nicefrac}       % compact symbols for 1/2, etc.
\usepackage{microtype}      % microtypography
\usepackage{lipsum}		% Can be removed after putting your text content
\usepackage{makecell} % for \makecell in tables
\usepackage{graphicx}
\usepackage{natbib}
\usepackage{doi}
\usepackage{lmodern}  % Better fonts than default Computer Modern
\usepackage{amsmath, amssymb}
\usepackage{graphicx}
\usepackage{booktabs}     % For nice tables
\usepackage{multirow}
\usepackage{float}
\usepackage{url}   % For clickable references
\usepackage{caption}
\usepackage{subcaption}
\usepackage{xspace}
\usepackage{siunitx}      % For number formatting
\usepackage{algpseudocode}
\usepackage{algorithm}
\usepackage{rotating}
\usepackage{array}
\usepackage{afterpage}
\hypersetup{
    colorlinks=false,
    pdfborder={0 0 0}
}
\usepackage{enumitem}

\newtheorem{df}{Definition}

\title{Dirichlet Process Mixtures within Semiparametric Bayesian Networks: A Novel Conditional Density Estimation Approach}

\author{
\href{https://orcid.org/0009-0000-2396-2446}{\includegraphics[scale=0.06]{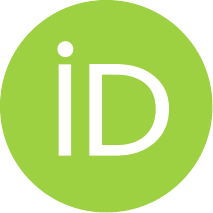}\hspace{1mm}Victor Alejandre} \\
Computational Intelligence Group \\
Departamento de Inteligencia Artificial \\
Universidad Politécnica de Madrid \\
Campus de Montegancedo, Boadilla del Monte \\
28660 Madrid, Spain \\
\texttt{victor.alejandre@upm.es} \\
\And
\href{https://orcid.org/0000-0003-0652-9872}{\includegraphics[scale=0.06]{orcid.pdf}\hspace{1mm}Pedro Larrañaga} \\
Computational Intelligence Group \\
Departamento de Inteligencia Artificial \\
Universidad Politécnica de Madrid \\
Campus de Montegancedo, Boadilla del Monte \\
28660 Madrid, Spain \\
\texttt{pedro.larranaga@fi.upm.es} \\
\And
\href{https://orcid.org/0000-0001-7109-2668}{\includegraphics[scale=0.06]{orcid.pdf}\hspace{1mm}Concha Bielza} \\
Computational Intelligence Group \\
Departamento de Inteligencia Artificial \\
Universidad Politécnica de Madrid \\
Campus de Montegancedo, Boadilla del Monte \\
28660 Madrid, Spain \\
\texttt{mcbielza@fi.upm.es} \\
}
\renewcommand{\shorttitle}{Dirichlet Process Mixtures within Semiparametric Bayesian Networks}

\hypersetup{
pdftitle={Dirichlet Process Mixtures within Semiparametric Bayesian Networks: A Novel Conditional Density Estimation Approach},
pdfsubject={CS, AI},
pdfauthor={Victor Alejandre, Concha Bielza, Pedro Larrañaga},
pdfkeywords={Semiparametric Bayesian networks, Multivariate kernel density estimation, Bandwidth selection, Cross-validation, Plug-in},
}

\begin{document}
\maketitle

\begin{abstract}
Semiparametric Bayesian networks (SPBNs) are flexible probabilistic graphical models for representing joint distributions of continuous variables through lower-dimensional conditional probability distributions (CPDs). Existing SPBN models use kernel density estimators (KDEs) for non-parametric CPD modeling, but their one-component-per-observation mixture representation can lead to increased inference costs and density estimation errors, which can be limiting. In this work, we define a new CPD modeling approach for non-parametric components based on Bayesian non-parametric Dirichlet Process Mixtures (DPMs) within SPBNs. DPMs provide a data-adaptive clustering of observations, yielding a more compact representation while retaining the flexibility of non-parametric density estimation, unlike KDE. This results in lower inference costs while improving density estimation performance. We implement the proposed DPM-based SPBN (DPM-SPBN) by extending the open-source \texttt{PyBNesian} package and evaluate it on synthetic and real-world data. For a given network structure, DPM-based CPDs significantly outperform state-of-the-art Bayesian network models for continuous variables in density estimation, achieving lower absolute log-likelihood errors on synthetic data. We further investigate the impact of the proposed CPD modeling on structure learning methods relying on CPD-based scores. DPM-based SPBNs achieve lower structure learning and density estimation errors than KDE-based SPBNs. Experiments on real-world data corroborate these findings, identifying DPM-based CPD modeling as the most effective alternative among the evaluated non-parametric approaches for continuous variables. 
\end{abstract}

% keywords can be removed
\keywords{Semiparametric Bayesian networks\and Dirichlet process mixtures\and Variational Bayes inference}

\section{Introduction}
\label{sec:intro}
A Bayesian network (BN) \cite{koller2009probabilistic} is a probabilistic graphical model designed to represent and learn the joint probability distribution (JPD) of a set of variables. The JPD provides a general probabilistic representation of the data-generating process, from which marginal and conditional distributions, as well as other quantities of interest, can be derived. Consequently, many statistical and machine learning problems, including exploratory data analysis, classification, regression, and probabilistic inference, can be formulated in terms of the JPD or properties derived from it. However, solving a predictive task does not necessarily require explicitly modeling the underlying probability distribution. Instead, many machine learning methods are designed to learn a predictive function or a specific functional of the underlying distribution. For example, under squared-error loss, the population-optimal regression function is the conditional expectation $\mathbb{E}[Y\mid X=x]$, which can be estimated without explicitly modeling the full conditional distribution or the JPD. Explicit probabilistic models, in contrast, characterize the underlying probability distribution and its associated uncertainty, from which predictive quantities can be derived, while also enabling a broader range of tasks, such as probabilistic inference and uncertainty quantification, that support informed decision-making.

Despite these advantages, probabilistic modeling introduces several practical challenges. First, tractable models often require strong assumptions about the underlying distribution, such as assuming a Gaussian distribution when the true data-generating process deviates from this family, potentially introducing model bias. Second, the number of parameters required to represent a JPD grows rapidly with the number of variables. For example, a categorical distribution over two binary variables requires three parameters, whereas representing the JPD of ten binary variables requires more than a thousand parameters. A similar issue arises when representing the covariance matrix of a multivariate Gaussian distribution. This growth increases both data requirements and computational complexity, making direct estimation increasingly impractical as the dimensionality increases.

BNs address these limitations by exploiting conditional independence relationships between variables. A BN decomposes the JPD into a set of lower-dimensional conditional probability distributions (CPDs), transforming the estimation of a JPD into a collection of more manageable local estimation problems. This factorization can substantially reduce the number of parameters that need to be estimated, thereby reducing computational costs and data requirements. Formally, a BN is defined by a directed acyclic graph (DAG), which encodes the conditional (in)dependence relationships among variables, together with a set of CPDs that characterize the corresponding local probabilistic relationships.

BNs have become a well-established framework for probabilistic modeling, with extensive research that addresses both their theoretical development and their application in diverse fields, including engineering, medicine and biology \cite{friedman2000using, bielza2014bayesian, WEBER2012671, kyrimi2021comprehensive}. The framework has been extended to accommodate discrete, continuous, and hybrid domains while preserving the underlying principles of probability theory. Accordingly, a wide range of parametric and non-parametric approaches have been developed for modeling CPDs. In parallel, substantial research has focused on learning the DAG structure from data, with the complexity and methodology of structure learning often depending on the variable types and the CPD models employed. Notable parametric extensions include Gaussian BNs \cite{gaussiannetworks} and conditional linear Gaussian BNs \cite{cgbnlaurizten}, which provide established formulations for continuous and hybrid domains, respectively.

For continuous variables, parametric CPD models can achieve faster convergence rates when the assumed distributional family adequately represents the underlying data generating process. However, misspecification of the assumed distribution can lead to substantial estimation errors. To address this limitation, several non-parametric approaches have been proposed, including Gaussian processes and kernel density estimators (KDEs). These methods avoid imposing a specific parametric distributional form and can therefore mitigate the bias associated with model misspecification, although this flexibility typically comes at the cost of slower convergence rates. In particular, KDEs achieve a mean integrated squared error (MISE) convergence rate of $O(N^{-4/(d+4)})$, where $N$ denotes the sample size and $d$ the dimensionality of the data, compared to the usual parametric rate of $O(N^{-1})$ (see Section 2.8 of \cite{chacon2018multivariate}).

Among the BN models developed for continuous and hybrid domains, semiparametric Bayesian networks (SPBNs) provide a flexible framework for CPD modeling \cite{atienza2022semiparametric, atienza2022hybrid}. SPBNs combine parametric and non-parametric approaches, allowing the CPD representation to be selected according to the characteristics of the data. In particular, SPBNs can employ linear Gaussian CPDs when the data are adequately described by the corresponding parametric assumptions, while KDEs provide a more flexible alternative when such assumptions are not appropriate. Thus, SPBNs combine the computational efficiency of parametric estimation with the flexibility of non-parametric estimation, providing an effective framework for modeling continuous and hybrid domains.

KDEs present several practical limitations when used for CPD modeling. A well-known challenge is the bias–variance trade-off, which has motivated extensive research on bandwidth selection methods to improve estimation accuracy \cite{alejandre2026bandwidth}. In addition, KDE-based models can become computationally demanding because the estimator typically contains one kernel component per observation, making density evaluation increasingly expensive as the sample size grows. Another important limitation is the curse of dimensionality, which is reflected in the convergence rate of the estimator and becomes increasingly pronounced as dimensionality $d$ increases. Although factorization provided by BNs reduces the dimensionality of individual CPD estimation problems, these limitations can still affect the accuracy and computational efficiency of CPD estimation in practice.

In this work, we propose a novel approach for nonparametric modeling of CPDs in SPBNs based on Dirichlet process mixture models (DPMs) \cite{antoniak1974mixtures}. By leveraging the clustering properties of DPMs, the proposed approach groups similar observations into a reduced number of mixture components, reducing the computational cost of density evaluation while retaining the flexibility of non-parametric modeling. As a Bayesian non-parametric mixture model, it provides a compact representation of CPDs that becomes particularly advantageous as the dimensionality $d$ of the data increases, where KDE-based approaches become increasingly challenging. We integrate this formulation into the SPBNs framework for the non-parametric nodes CPD estimation to obtain DPM-SPBNs and conduct extensive experiments on synthetic and real-world data. We compare DPM-SPBNs with KDE-based SPBNs and state-of-the-art BN models for continuous variables in terms of density estimation and computational requirements. We further investigate the effect of the proposed CPD formulation on structure learning by combining DPM-based CPDs with hill-climbing (HC) and comparing the resulting DPM-SPBNs with their KDE-based counterparts. Overall, the results demonstrate improved density estimation, substantially lower inference costs, and moderate training times, establishing DPM-SPBNs as a leading alternative for non-parametric modeling of continuous variables in BNs.

The remainder of the paper is organized as follows. Section \ref{sec:2} reviews the theoretical background on DPMs and BNs. Section \ref{sec:3} presents and develops the proposed DPM-SPBN framework. Section \ref{sec:4} describes the experimental evaluation, examining the impact of DPM-based CPD modeling on learning of SPBN parameters and structures. Finally, Section \ref{sec:5} concludes the paper and outlines the directions for future research.

\section{Background}
\label{sec:2}
\subsection{Dirichlet Process Mixture}
\label{sec:dpm}
The Dirichlet process (DP) \cite{ferguson1973bayesian} is a probability measure over probability measures, parameterized by a concentration parameter $\alpha > 0$ and a base probability measure $\mathbf{G}_0$. The concentration parameter $\alpha$ controls the concentration of the process around the base measure $\mathbf{G}_0$, which determines the location of the process in the space of probability measures. We denote a random probability measure $\mathbf{G}$ following a DP with parameters $\alpha$ and $\mathbf{G}_0$ as $\mathbf{G} \sim \mathrm{DP}(\alpha,\mathbf{G}_0).$

Formally, let $\Theta$ denote a measurable space and let $\mathbf{G}_0$ be a probability measure defined on $\Theta$. For any finite measurable partition ${B_1,\ldots,B_k}$ of $\Theta$, the probabilities assigned by $\mathbf{G}$ to the elements of the partition satisfy
\begin{equation}
(\mathbf{G}(B_1),\ldots,\mathbf{G}(B_k))
\sim
\mathrm{Dir}\left(
\alpha\mathbf{G}_0(B_1),\ldots,\alpha\mathbf{G}_0(B_k)
\right).
\end{equation}
where $\mathrm{Dir}\left(\cdot,...,\cdot\right)$ represents the Dirichlet distribution. A key property of the DP is that a draw $\mathbf{G}$ is discrete with probability one, even when the base measure $\mathbf{G}_0$ is continuous. Consequently, samples drawn from $\mathbf{G}$ can take identical values, resulting in a finite number of distinct values for any finite sample. This property provides the basis for the construction of the non-parametric Dirichlet process mixture (DPM) model \cite{antoniak1974mixtures} presented next.

Let $\mathbf{X}$ denote a generic random vector taking values in an observation space $\mathcal{X}$, and let $P(\mathbf{x}\mid\theta)$ denote a family of probability distributions on $\mathcal{X}$ indexed by a parameter $\theta\in\Theta$. In the usual sampling setting, a random sample $\mathbf{X}^1,\ldots,\mathbf{X}^N$ is generated from a common distribution for $\mathbf{X}$. In the DPM, we represent this common distribution hierarchically by introducing a latent parameter for each observation. Specifically, rather than fixing a single parameter $\theta$ for the distribution of $\mathbf{X}$, we allow each observation $\mathbf{X}^i$ to have its own latent parameter $\theta^i$, with the parameter drawn from a common random probability measure $\mathbf{G} \sim \mathrm{DP}(\alpha,\mathbf{G}_0)$. Conditional on $\theta^i$, the observation is generated according to

$$\mathbf{X}^i\mid\theta^i\sim P(\cdot\mid\theta^i).$$

Thus, the individual parameters \(\theta^i\) provide a latent representation of how each observation is generated, while the common random measure $\mathbf{G}$ determines how these parameters are distributed across observations. For a sample of random vectors $\mathbf{X}^1,\ldots,\mathbf{X}^N$, the hierarchical model resulting is therefore given by
\begin{equation}
\mathbf{G} \mid \alpha,\mathbf{G}_0 \sim \mathrm{DP}(\alpha,\mathbf{G}_0),
\qquad
\theta^i \mid \mathbf{G} \sim \mathbf{G},
\qquad
\mathbf{X}^i\mid\theta^i\sim P(\cdot\mid\theta^i),
\qquad i=1,\ldots,N.
\end{equation}
The hierarchy separates the generation of each observation into two stages: first, a latent parameter $\theta^i$ is drawn from the same random measure $\mathbf{G}$ for every observation, and second, $\mathbf{X}^i$ is generated from the distribution indexed by that parameter. Hence, conditional on $\mathbf{G}$, the latent parameters $\theta^1,\ldots,\theta^N$ are independent and identically distributed according to $\mathbf{G}$. Integrating out the individual latent parameter $\theta^i$ then gives the common distribution of the observations conditional on $\mathbf{G}$,
\begin{equation}
\mathbf{X}^i\mid\mathbf{G}
\overset{\mathrm{iid}}{\sim}
P(\mathbf{x}\mid\mathbf{G}) = \int P(\mathbf{x}\mid\theta)d\mathbf{G}(\theta).
\end{equation}

The hierarchical representation provides a construction of the common distribution through the latent parameters $\theta^i$ and the shared random measure $\mathbf{G}$. Since a draw $\mathbf{G}$ from a DP is discrete, different observations can draw the same parameter value, i.e., $\theta^i=\theta^j$ for $i\neq j$. Such observations are therefore generated from the same component distribution, effectively grouping these data points into a shared cluster. Consequently, the common sampling distribution $P(\mathbf{x}\mid\mathbf{G})$ takes the form of a mixture distribution, with an unbounded number of potential components but only a finite number of occupied components for any finite dataset.

The mixture structure induced by the DP can be made explicit through its stick-breaking representation \cite{sethuraman1994constructive}. Let $\{V_k\}_{k=1}^{\infty}$ be independent random variables with $V_k\sim\mathrm{Beta}(1,\alpha)$, and let $\{\tilde{\theta}^k\}_{k=1}^{\infty}$ be independent draws from the base measure $\mathbf{G}_0$. The random measure $\mathbf{G}$ can then be represented as
$$
\pi_k = V_k\prod_{j=1}^{k-1}(1-V_j), \qquad
\mathbf{G} = \sum_{k=1}^{\infty}\pi_k\delta_{\Tilde{\theta}^k},
$$
where $\pi_k$ denotes the weight associated with the $k$-th atom and $\delta_{\tilde{\theta}^k}$ is a Dirac measure centered at $\tilde{\theta}^k$. Hence, $\mathbf{G}$ consists of a countably infinite collection of potential mixture components with weights $\pi_k$ and component parameters $\tilde{\theta}^k$. For a fixed realization of $\mathbf{G}$, the marginal density of an observation is obtained by integrating out its latent parameter:
\begin{equation}
P(\mathbf{x}\mid\mathbf{G})=\int P(\mathbf{x}\mid\theta)d\mathbf{G}(\theta)=\sum_{k=1}^{\infty}\pi_k P(\mathbf{x}\mid\tilde{\theta}^k).
\end{equation}

Thus, conditional on $\mathbf{G}$, the distribution of $\mathbf{X}$ is an infinite mixture. However, $\mathbf{G}$ itself is random and, therefore, so is the resulting mixture distribution. To obtain a marginal or predictive distribution that does not condition on a particular realization of $\mathbf{G}$, the uncertainty associated with $\mathbf{G}$ must be integrated out.

Before observing any data, this marginalization is particularly simple. Since the expectation of a DP draw is its base measure ($\mathbb{E}[\mathbf{G}]=\mathbf{G}_0$), the predictive density of an observation is
\begin{equation}
P(\mathbf{x}\mid\alpha,\mathbf{G}_0)=\int P(\mathbf{x}\mid\theta)d\mathbf{G}_0(\theta).
\end{equation}

However, once observations are available, integrating out $\mathbf{G}$ no longer results in independent draws of the latent parameters. Assume that we observe a realization $\mathbf{x}^1, \dots, \mathbf{x}^N$ of the random sample $\mathbf{X}^1, \dots, \mathbf{X}^N$. Under the generative hierarchy of the DPM, each observation $\mathbf{x}^i$ is governed by a corresponding latent parameter $\theta^i$, with all parameters $\theta^1, \dots, \theta^N$ sampled from the same random measure $\mathbf{G}$. Because the latent parameters are drawn from a shared distribution, marginalizing out $\mathbf{G}$ induces statistical dependencies among them, collapsing the infinite-dimensional prior into a marginal JPD over the sequence of parameters. This dependence can be characterized sequentially through the Pólya urn scheme \cite{blackwell1973ferguson}, which defines the conditional distribution of a new parameter $\theta^{N+1}$ given the previously drawn parameters:
$$P(\theta^{N+1} \mid \theta^1, \dots, \theta^N) = \frac{\alpha}{\alpha + N} \mathbf{G}_0(\theta^{N+1}) + \frac{1}{\alpha + N} \sum_{i=1}^{N} \delta_{\theta^i}(\theta^{N+1}).$$

Hence, a new parameter is drawn from either the base measure or from one of the previously instantiated parameter values, with probabilities determined by $\alpha$ and the number of previous observations. Let $\{\theta_1^*, \dots, \theta_C^*\}$ denote the unique values among the parameters $\{\theta^1, \dots, \theta^N\}$. These unique values represent the active mixture components from which the data have been sampled, where observations that share the same parameter group together to form $C$ distinct clusters. The conditional distribution of $\theta^{N+1}$ takes the following form:
$$
\theta^{N+1} = 
\begin{cases} 
\theta_c^* & \text{with probability } \frac{n_c}{\alpha + N}, c=1,...,C \\ 
\theta \sim \mathbf{G}_0 & \text{with probability } \frac{\alpha}{\alpha + N}, 
\end{cases} 
$$
where $n_c$ is the number of observations assigned to cluster $c$. 

Building upon this parameter-level distribution, we can derive the conditional predictive density for a new observation. If we assume that for a given realization of the data sample $\mathbf{x}^1, \dots, \mathbf{x}^N$, the associated latent parameters $\theta^1, \dots, \theta^N$ are known, the predictive density of a new point $\mathbf{x}^{N+1}$ evaluates as follows:
\begin{equation}
P(\mathbf{x}^{N+1} \mid \theta^1, \dots, \theta^N, \alpha, \mathbf{G}_0) = \frac{\alpha}{\alpha + N} \int P(\mathbf{x}^{N+1} \mid \theta) \, d\mathbf{G}_0(\theta) + \sum_{c=1}^{C} \frac{n_c}{\alpha + N} \, P(\mathbf{x}^{N+1} \mid \theta_c^*).
\label{eq:predictive_density}
\end{equation}

\begin{figure}[htbp]
    \centering
    \includegraphics[width=1\textwidth]{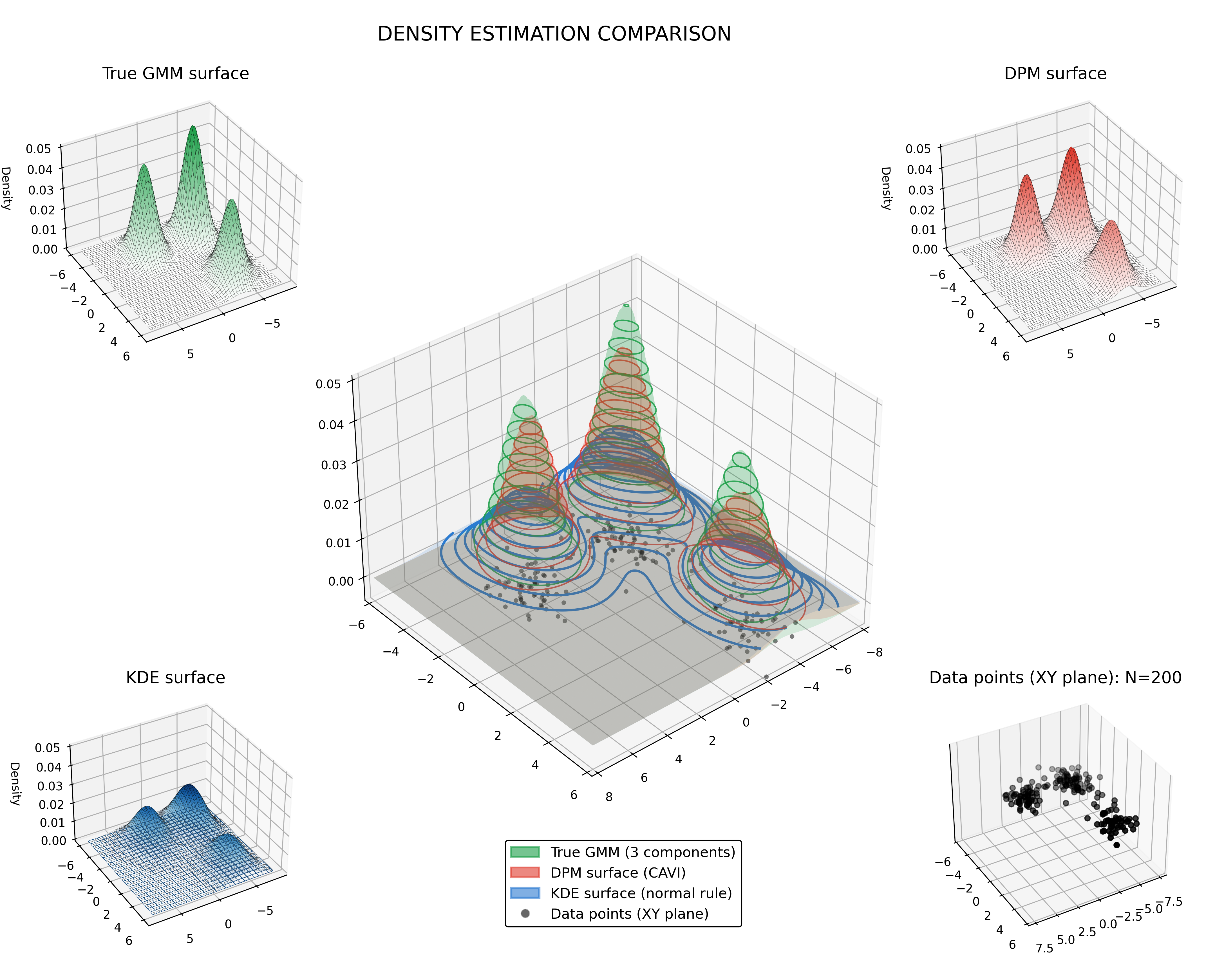}
    \caption{Comparison of density estimation using a DPM and a KDE with a $200$-sample from a true underlying three-component Gaussian mixture.}
    \label{fig:density_comparison}
\end{figure}

This clustering mechanism provides an important advantage over non-parametric approaches such as KDEs, which associate one kernel component with each observation. In contrast, DPMs share latent parameters among observations assigned to the same cluster/component, resulting in a more compact representation with $C \ll N$. This reduces the computational cost of density evaluation while retaining the flexibility of non-parametric density estimation. Moreover, as dimensionality increases, observations become increasingly sparse in the feature space, making the one-component-per-observation representation of KDEs less informative and increasing the amount of data required for accurate estimation. By sharing parameters across similar observations, DPMs concentrate the density representation on occupied regions of the space, potentially mitigating these effects and providing more accurate density estimates in higher dimensional settings (see Figure \ref{fig:density_comparison} for an example).

However, the latent cluster parameters are not observed; only the data sample $\mathbf{x}^1,\dots,\mathbf{x}^N$ is available. Therefore, obtaining the predictive density requires marginalizing over the sequence of latent parameters $\boldsymbol{\theta}_{1:N} = (\theta^1, \dots, \theta^N)$:

\begin{equation}P(\mathbf{x}^{N+1} \mid \mathcal{D}) = \int P(\mathbf{x}^{N+1} \mid \boldsymbol{\theta}_{1:N}) P(\boldsymbol{\theta}_{1:N} \mid \mathcal{D}) , d\boldsymbol{\theta}_{1:N}.
\end{equation}

Because $\mathbf{G}$ has been integrated out, the latent parameters become dependent. Since the shared measure $\mathbf{G}$ is discrete, different observations have a real probability of sharing the exact same parameter value ($P(\theta^i = \theta^j) > 0$), naturally grouping the data into clusters. Due to these exact discrete ties, evaluating the predictive density through standard continuous integration is mathematically impossible.

To resolve this, we change the representation of the parameter space. The sequence $\boldsymbol{\theta}_{1:N}$ is bijectively mapped to two elements: a latent partition variable $\mathbf{c}$ defining the cluster assignments, and a vector of unique continuous parameters $\boldsymbol{\theta}^* = (\theta_1^*, \dots, \theta_C^*)$ corresponding to these distinct clusters. This bijection ensures that their posterior probability distributions represent the equivalent latent state, yielding $P(\boldsymbol{\theta}_{1:N} \mid \mathcal{D}) \equiv P(\mathbf{c}, \boldsymbol{\theta}^* \mid \mathcal{D})$. This transformation provides a mathematically tractable representation of the original parameter sequence.

Under this restructured hierarchy, the probabilities of these clustering partitions are governed by the Pólya urn scheme. Before observing any data, marginalizing out $\mathbf{G}$ yields the prior probability of any given partition $P(\mathbf{c})$. However, once the dataset $\mathcal{D}$ is available, integrating out $\mathbf{G}$ in the presence of the data likelihood directly yields the joint posterior distribution on both the discrete partitions and their unique parameters, $P(\mathbf{c}, \boldsymbol{\theta}^* \mid \mathcal{D})$.

Because the actual partition and the continuous cluster parameters remain fundamentally unobserved, fully Bayesian inference dictates that we cannot condition on any single clustering arrangement. Since we only have access to the observed dataset $\mathcal{D}$, we still have to account for our uncertainty regarding these hidden variables by marginalizing over both levels of the latent hierarchy, having already integrated out $\mathbf{G}$. Specifically, this involves taking the expectation of the conditional data likelihood $P(\mathbf{x}^{N+1} \mid \mathbf{c}, \boldsymbol{\theta}^*)$ over the joint posterior $P(\mathbf{c}, \boldsymbol{\theta}^* \mid \mathcal{D})$. By applying the chain rule of probability, this expectation naturally decomposes: it requires summing over all possible discrete partitions, weighted by their marginal posterior probability $P(\mathbf{c} \mid \mathcal{D})$, and integrating out the corresponding unique continuous cluster parameters according to their conditional posterior distribution $P(\boldsymbol{\theta}^* \mid \mathbf{c}, \mathcal{D})$:

\begin{equation}
P(\mathbf{x}^{N+1} \mid \mathcal{D}) = \sum_{\mathbf{c}} P(\mathbf{c} \mid \mathcal{D}) \int P(\mathbf{x}^{N+1} \mid \mathbf{c}, \boldsymbol{\theta}^*) P(\boldsymbol{\theta}^* \mid \mathbf{c}, \mathcal{D}) \, d\boldsymbol{\theta}^*
\end{equation}

For a fixed partition $\mathbf{c}$, the dataset is divided into independent clusters. Under this conditioned state, the base measure $\mathbf{G}_0$ serves as the exact independent prior for each unique continuous parameter $\theta_c^*$. Consequently, the conditional posterior $P(\boldsymbol{\theta}^* \mid \mathbf{c}, \mathcal{D})$ factorizes into independent updates for each cluster. If a conjugate base distribution $\mathbf{G}_0$ is used, this inner integral over the continuous parameters can be resolved analytically.

However, the outer summation requires evaluating every possible arrangement of $N$ observations in any number of clusters. The total number of valid partitions is given by the Bell number, which grows super-exponentially with the dataset size. This combinatorial explosion makes the exact marginalization fundamentally intractable. Consequently, DPM inference must rely on approximate methods to characterize the posterior distribution over the latent variables and estimate the predictive density.

Markov Chain Monte Carlo (MCMC) is widely used for posterior inference over partitions and mixture parameters in DPM models. Early methods marginalized the random measure $\mathbf{G}$. Because $\mathbf{G}$ is composed of infinite mixture weights and cluster parameters, this marginalization has a dual effect: it replaces the infinite weights with discrete cluster assignments $\mathbf{c}$, while leaving the unique continuous cluster parameters $\boldsymbol{\theta}^*$ governed by the base measure. In conjugate models, these remaining continuous parameters can then be integrated out analytically, yielding a fully collapsed Gibbs sampler \cite{Maceachern1994}. The resulting Markov chain contains only the discrete cluster assignments, substantially simplifying inference. However, in non-conjugate models, assigning an observation to a new cluster requires the marginal likelihood which is generally intractable. Auxiliary variable methods avoid this integral by introducing explicit parameters for potential new clusters \cite{maceachern1998estimating}. Alternatively, Metropolis-Hastings methods propose new cluster parameters from the prior  \cite{neal2000markov}. Subsequent advances include blocked Gibbs samplers based on truncated stick-breaking representations \cite{ishwaran2001gibbs}, split-merge moves for improved exploration of the partition space \cite{jain2004split}, and alternative approaches such as slice sampling \cite{kalli2011slice,hastie2015sampling} and sequential Monte Carlo \cite{ulker2010sequential}. Despite these developments, MCMC remains computationally expensive and can converge slowly, motivating faster approximate inference methods.

Variational inference (VI) provides a scalable framework for approximate inference in DPMs. Early approaches developed mean-field variational algorithms for conjugate exponential-family DPMs using truncated stick-breaking representations \cite{jordan_blei}, while related work investigated collapsed variational inference, which analytically integrates out the mixture weights \cite{kurihara2007collapsed}. To improve scalability for large datasets, stochastic variational inference (SVI) introduced mini-batch updates \cite{hoffman}, which were subsequently adapted to DPMs \cite{wang2012truncation, hughes2013memoized, lin2013online}. However, despite being described as online methods, these approaches still require the total dataset size $N$ to scale their updates. Subsequent approaches have addressed this limitation by enabling continual inference without dependence on a predefined $N$ or fixed truncation, using sequential recursive Bayesian updating \cite{huynh2016streaming, yang2019memorized} or asynchronous posterior decomposition with combinatorial matching \cite{campbell2015streaming}.

Finally, another class of approximate inference methods relies on hard-clustering and greedy strategies that avoid full posterior integration. These approaches typically select a single partition, or construct an approximation concentrated around a limited set of partitions, making inference computationally tractable at the cost of reduced posterior uncertainty. Prominent examples include sequential updating and greedy search (SUGS) \cite{wang2011fast} and its variational refinement \cite{Zhang02102014}, which perform sequential updates to obtain efficient approximate Bayesian estimates. Related approaches based on maximum a posteriori, such as MAP-DP \cite{raykov}, provide computationally efficient maximum a posteriori inference for exponential-family DPMs. Overall, these methods trade full posterior averaging for substantial computational gains.

\subsection{Bayesian networks}
\label{sec:bayesian_networks}
A BN \cite{koller2009probabilistic} is a probabilistic graphical model that provides a compact representation of a JPD by exploiting conditional independence relationships among its random variables. The formal definitions follow.

\begin{df}
Let $\mathbf{X}=(X_{1},\ldots,X_{d})$ be a random vector with JPD $P(X_{1},\ldots,X_{d})$. A BN is defined as the tuple $\mathcal{B}=(\mathcal{G},\boldsymbol{\theta})$, where $\mathcal{G}=(\mathbf{V},\mathbf{A})$ is a \textit{directed acyclic graph} (DAG). The set of vertices is $\mathbf{V}=\{X_{1},\ldots,X_{d}\}$, and $\mathbf{A}\subseteq\mathbf{V}\times\mathbf{V}$ denotes the set of directed arcs. The parameter set $\boldsymbol{\theta}=\{P(X_j\mid\mathbf{X}_{Pa(j)}),j=1,\ldots,d\}$ specifies the CPD associated with each variable $X_j$ given its parent variables $\mathbf{X}_{Pa(j)}$ in $\mathcal{G}$, where $Pa(j)$ denotes the set of indices corresponding to the parents of $X_j$.
\label{definition2:3}
\end{df}

The resulting factorization of $P(X_{1},...,X_{d})$ is:
\[P(X_{1},...,X_{d})=\prod_{j=1}^dP(X_{j}\mid \mathbf{X}_{Pa(j)}).\]
The factorization induced by the DAG decomposes the JPD into a set of local CPDs, making their estimation more tractable. Both the DAG structure and the corresponding CPDs can be learned from data, a problem that has been extensively studied in the literature. In addition, the combination of the graphical structure and associated probability distributions provides an interpretable representation of the dependencies among variables. Together with the different types of probabilistic inference that can be performed within the model, this representation supports a wide range of approaches for interpreting and reasoning about the relationships encoded by the BN.

\begin{figure}[htbp]
    \centering
    \includegraphics[width=0.8\textwidth]{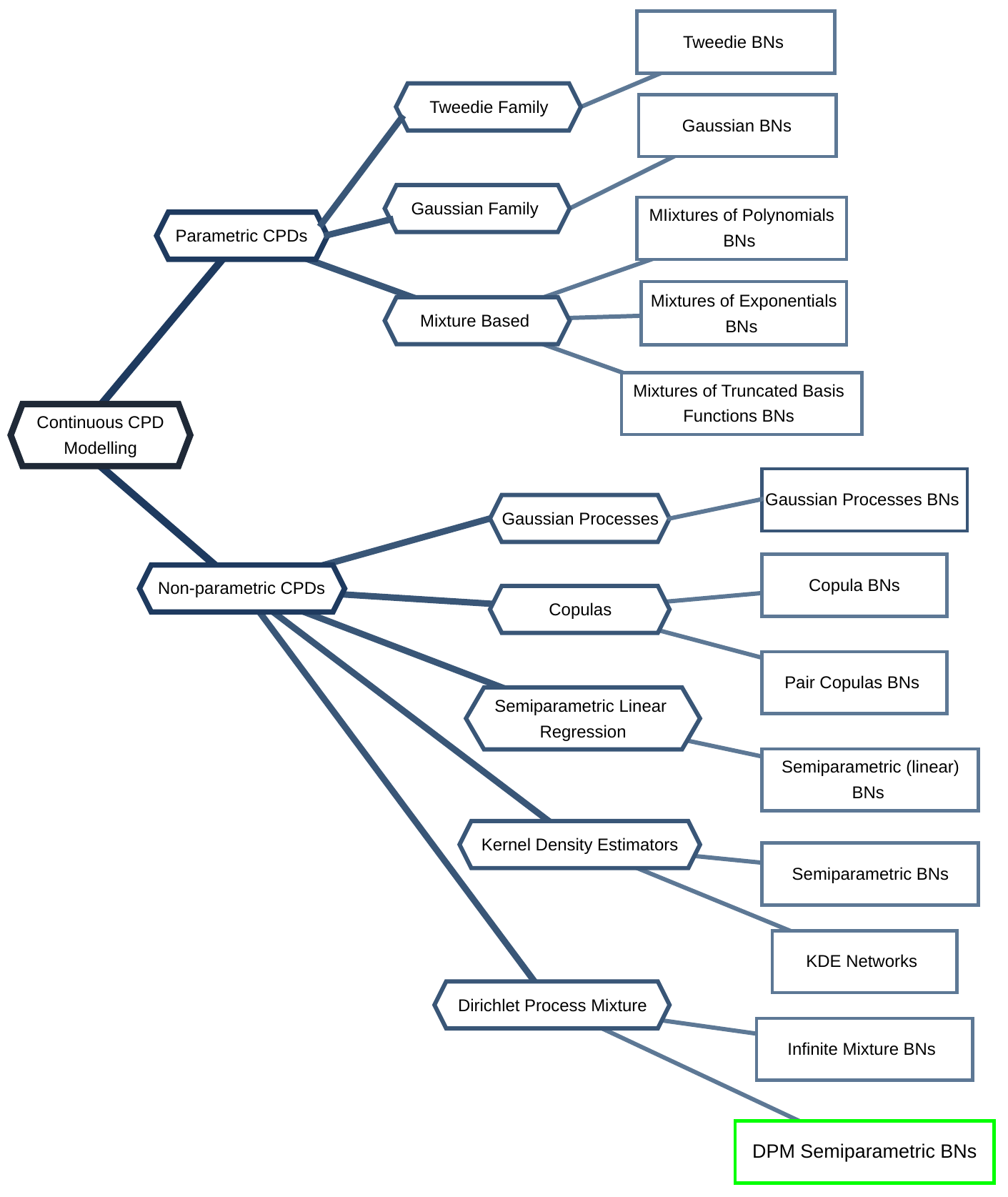}
    \caption{Schematic overview of parametric and non-parametric approaches for modeling continuous CPDs in BNs. Each approach branches into different models and underlying assumptions, with the resulting BNs shown in rectangles. The novel proposal introduced in this work is highlighted in green.}
    \label{fig:continous CPD}
\end{figure}

BNs can employ either parametric or non-parametric CPDs. In the following, given the scope of this work, we focus on CPD modeling for continuous variables (see Figure \ref{fig:continous CPD}), although BNs can also accommodate other variable types, such as discrete variables with categorical CPDs. Common parametric approaches include: (i) Gaussian BNs, which assume a multivariate Gaussian JPD and use linear Gaussian (LG) CPDs \cite{gaussiannetworks}; (ii) Tweedie BNs, whose CPDs belong to the Tweedie family, a subclass of the exponential family \cite{masmoudi2019new}; and (iii) mixture-based models, including mixtures of truncated exponentials \cite{moral2001mixtures}, polynomials \cite{shenoy2011inference}, and truncated basis functions \cite{langseth2012mixtures}.

Non-parametric CPD models relax these distributional assumptions and include Gaussian processes \cite{Friedman2000GaussianPN}, infinite mixtures \cite{ickstadt2010nonparametric}, and kernel density estimators (KDEs) \cite{atienza2022semiparametric, hofmann1995discovering}. These approaches provide greater flexibility, but can incur higher computational costs, particularly as the sample size increases. Gaussian process  BNs can become computationally demanding with increasing sample size. Infinite-mixture models were introduced into BNs by \cite{ickstadt2010nonparametric}, who combined DPMs with LG CPDs to obtain a non-parametric CPD representation. However, their formulation was developed exclusively for BN structure learning: the mixture-component parameters were integrated out, and the resulting model was not intended to provide estimates of the final CPDs. To the best of our knowledge, this formulation has not been further developed for posterior inference of the CPDs. Moreover, the structure-learning procedure is based on computationally intensive MCMC sampling. KDEs avoid some of these costs, but remain subject to the bias-variance trade-off, the curse of dimensionality, and increasing density evaluation costs with the sample size. In \cite{Boukabour17122021}, a semiparametric approach is proposed, which models each CPD using a partially linear semiparametric regression.

Copula-based models provide another flexible approach to non-parametric CPD modeling. Pair-copula BNs \cite{kurowicka2005distribution} use bivariate conditional copulas to model dependencies while allowing non-parametric estimation of the marginals. However, general pair-copula constructions can be computationally demanding for density evaluation and simulation, whereas Gaussian copulas offer greater tractability at the cost of potentially restrictive dependence assumptions. Alternatively, local multivariate copulas can be used to directly parameterize CPDs \cite{elidan2010copula}, although their density evaluation may require costly computation of copula derivatives given copula choice. Overall, non-parametric and semiparametric approaches provide greater flexibility than conventional parametric CPDs, but typically introduce additional computational or modeling trade-offs.

Structure learning \cite{kitson2023survey} concerns the recovery of the DAG from data and constitutes a broad and active area of research. Existing approaches are commonly categorized into three main classes: (i) constraint-based methods, which identify conditional independencies using statistical tests and construct a graph consistent with the detected relationships; (ii) score-based methods, which search for a graph that optimizes a statistical score measuring its fit to the data; and (iii) hybrid methods, which combine elements of both approaches. For score-based methods, a scoring criterion and a search procedure are required, with HC combined with Bayesian scores and its variants being among the most widely used approaches \cite{cooper1992bayesian}.

Structure learning methods are often closely related to the assumptions used for CPD modeling. For example, Bayesian scores may rely on Gaussian or categorical CPDs. Similarly, constraint-based methods may require conditional independence tests whose validity depends on the underlying assumptions, with tests based on Gaussian  being common. Consequently, the choice of the CPD model can affect not only parameter estimation but also the computational cost and feasibility of structure learning.

Several of the BN models discussed above have therefore been considered together with specific structure learning procedures. The non-parametric BN of \cite{ickstadt2010nonparametric} uses MCMC to recover the structure while marginalizing the model parameters, making it primarily a structure learning methodology. For pair-copula BNs, Gaussian copulas enable computationally efficient structure learning  using rank correlations \cite{hanea2015non}, while general non-Gaussian pair-copula models typically require computationally expensive numerical integration, depend on a prescribed parent ordering, and rely on adapted structure estimation methods such as HC with AIC/BIC  scores or a modified PC algorithm \cite{bauer2016pair}. For CPDs based on local multivariate copulas \cite{elidan2010copula}, both score-based and constraint-based structure learning approaches have been proposed, including continuous PC, which uses non-parametric conditional independence tests \cite{lasserre2021constraint}. The semiparametric BN of \cite{Boukabour17122021} uses sequential hypothesis testing, specifically Student's t-tests for linear links and bootstrapping non-parametric tests for nonlinear links based on a predetermined topological ordering.

In general, CPD modeling and structure learning constitute distinct but closely related problems, as the choice of CPD formulation can affect the assumptions and computational procedures underlying structure learning.

\section{DPM-SPBN}
\label{sec:3}
\begin{figure}[htbp]
    \centering
    \includegraphics[width=1\textwidth]{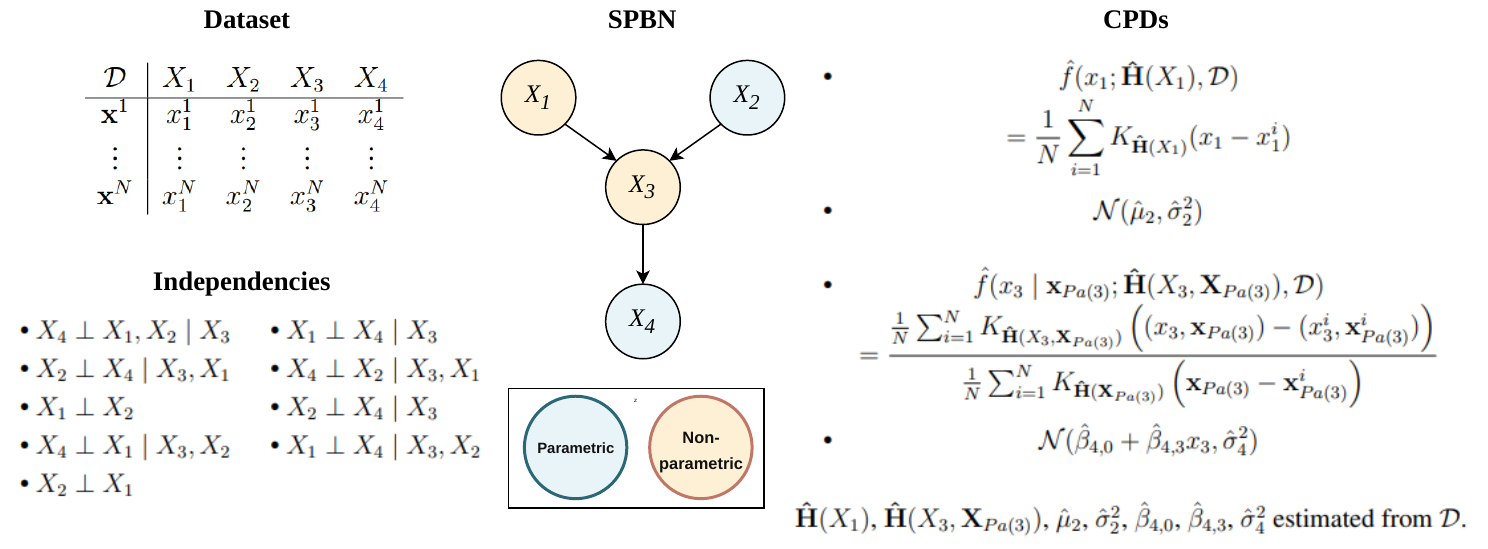}
    \caption{Example of a SPBN for JPD estimation. The figure illustrates a sample dataset $\mathcal{D}$, the network structure distinguishing non-parametric ($X_1, X_3$) and parametric ($X_2, X_4$) nodes, the estimated CPDs derived from $\mathcal{D}$ and encoded conditional independencies.} 
    \label{fig:spbn_example}
\end{figure}

In this section, we introduce the novel non-parametric CPD formulation based on DPMs within the SPBN framework. The semiparametric nature of SPBNs arises from combining different CPD formulations within the same network, with each CPD modeled  parametrically or non-parametrically (see Figure \ref{fig:spbn_example} for an example). The proposed DPM-SPBN replaces the existing KDE-based formulation for non-parametric CPDs with the proposed DPM-based formulation.

\subsection{CPD learning}
For the parametric component, the original SPBN formulation employs LG CPDs \cite{gaussiannetworks}, as established in Gaussian BNs. These CPDs model the conditional mean of a variable as a linear function of its parents, with Gaussian conditional errors. The proposed DPM-SPBN retains this formulation for parametric CPDs.

The non-parametric component employs KDEs, specifically conditional kernel density estimators (CKDEs), to model the CPDs. Their formal definition is given below.

\begin{df}
\label{definition3:1}
Let $X_j$ be a random variable, and $\mathbf{X}_{Pa(j)}$ be its random vector of parents of dimension $|Pa(j)|$. Given a realized dataset of $N$ i.i.d. observations $\mathcal{D} = \{(x_j^1, \mathbf{x}_{Pa(j)}^1), \dots, (x_j^N, \mathbf{x}_{Pa(j)}^N)\}$, the CKDE estimator of the conditional density of $X_j$ given $\mathbf{X}_{Pa(j)}$ is defined as:
\begin{equation}
\hat{f}(x_j \mid \mathbf{x}_{Pa(j)}; \mathbf{H}, \mathcal{D}) = \frac{\hat{f}(x_j, \mathbf{x}_{Pa(j)}; \mathbf{H}(X_j, \mathbf{X}_{Pa(j)}), \mathcal{D})}{\hat{f}(\mathbf{x}_{Pa(j)}; \mathbf{H}(\mathbf{X}_{Pa(j)}), \mathcal{D})} = \frac{\frac{1}{N} \sum_{i=1}^N K_{\mathbf{H}(X_j, \mathbf{X}_{Pa(j)})}\left((x_j, \mathbf{x}_{Pa(j)}) - (x_j^i, \mathbf{x}_{Pa(j)}^i)\right)}{\frac{1}{N} \sum_{i=1}^N K_{\mathbf{H}(\mathbf{X}_{Pa(j)})}\left(\mathbf{x}_{Pa(j)} - \mathbf{x}_{Pa(j)}^i\right)}.
\label{eq:CKDE}
\end{equation}

Here, $K_{\mathbf{H}}$ denotes the multivariate Gaussian kernel, and the bandwidth matrices $\mathbf{H}(X_j, \mathbf{X}_{Pa(j)})$ and $\mathbf{H}(\mathbf{X}_{Pa(j)})$ are deterministically coupled such that:
\begin{equation}
\begin{aligned}
\mathbf{H}(X_j, \mathbf{X}_{Pa(j)}) &= \begin{bmatrix}
a & \mathbf{v}^{\mathsf{T}} \\
\mathbf{v} & \mathbf{M}
\end{bmatrix}, \text{where} \quad a\in\mathbb{R}, \hspace{1mm}\mathbf{v}\in \mathbb{R}^{|Pa(j)|},\text{ and }\mathbf{M}\in\mathbb{M}^{|Pa(j)|\times |Pa(j)|} \text{ is a principal submatrix.}\\
\mathbf{H}(\mathbf{X}_{Pa(j)}) &= \mathbf{M}
\end{aligned}
\label{eq:bandwidth_relationship}
\end{equation}
\end{df}

Bandwidth matrix selection is only needed for $\mathbf{H}(X_j,\mathbf{X}_{Pa(j)})$ given the relationship in Equation \ref{eq:bandwidth_relationship}. It was originally selected using the normal reference rule (NR) \cite{Wand1992}, whose normality assumption can lead to oversmoothing and underfitting when it is violated. To address this limitation, \cite{alejandre2026bandwidth} proposed cross-validation and plug-in methods for bandwidth selection, highlighting the performance of unbiased cross-validation (UCV). These methods provide a more favorable bias-variance trade-off and more accurate estimates, at the cost of increased computational complexity.

Despite these improvements, KDEs remain subject to two fundamental limitations. First, the curse of dimensionality can substantially degrade estimation accuracy, as reflected by the MISE convergence rate (see Section \ref{sec:intro}). Second, the computational cost of density evaluation increases with the sample size, since the KDE requires a summation over all observations (Definition \ref{definition3:1}). We next introduce the non-parametric CPD formulation based on DPMs, which addresses the computational limitations of CKDEs through their inherent clustering structure, as detailed in Section \ref{sec:dpm}.  Before defining the proposed conditional estimator, we first discuss the two key considerations that motivate its formulation.

The first issue concerns posterior inference for DPM given a dataset \(\mathcal{D}\). As discussed in Section \ref{sec:dpm}, the DPM posterior is generally intractable when cluster assignments are unknown, motivating approximate inference methods such as MCMC, variational inference, and hard-clustering approaches.

MCMC methods approximate the posterior predictive density by averaging over the retained samples $B$ from the posterior state space, which can introduce substantial computational overhead. This can be illustrated using a fully collapsed Gibbs sampler for a conjugate base distribution, in which the component parameters are analytically marginalized and the sampler generates only a sequence of partitions $\mathbf{c}^{(b)}$. The posterior predictive density for a new observation $\mathbf{x}$ given the observed data $\mathcal{D}$ is then estimated as
\begin{equation}
\hat{f}(\mathbf{x}\mid\mathcal{D},\alpha,\boldsymbol{\eta})
\approx
\frac{1}{B}
\sum_{b=1}^{B}
P(\mathbf{x}\mid\mathbf{c}^{(b)},\mathcal{D},\alpha,\boldsymbol{\eta}),
\end{equation}
where $\boldsymbol{\eta}$ denotes the hyperparameters of the base distribution. Under this conjugate formulation, each sampled partition $\mathbf{c}^{(b)}$ determines the cluster assignments used to compute the corresponding posterior parameters. For a Gaussian likelihood with a Normal-Wishart (NW) base distribution, these posterior parameters define a multivariate Student's $t$-distribution for the conditional predictive density associated with each partition. Although each predictive density can be evaluated in closed form, MCMC-based prediction requires computing it for every retained posterior state sampled and averaging the resulting densities. This repeated evaluation can become computationally costly as the number of retained samples increases for convergence.

Although each sampled partition may contain substantially fewer components than the $N$ kernels of a CKDE, the need to account for multiple posterior states can offset the computational benefits of the more compact DPM representation. Furthermore, the cost of MCMC inference is incurred during the CPD learning itself, rather than only at the prediction time. This learning phase is highly expensive because the algorithm must evaluate the likelihood of every single observation against all active clusters across thousands of iterations to achieve convergence and collect valid samples. This computational burden can become particularly restrictive for the learning of multiple CPDs. 

Hard-clustering methods such as SUGS avoid MCMC sampling by greedily assigning the observations to a single fixed partition. The resulting predictive density follows the finite-mixture formulation in Equation \ref{eq:predictive_density}, conditioned on the fixed partition and the corresponding component parameters. However, in a conjugate setting, the component parameters can be analytically integrated out, so the predictive density is expressed directly in terms of the observations assigned to each cluster. For a multivariate Gaussian likelihood with a NW base distribution, these integrals yield multivariate Student's $t$-distributions:

\begin{equation}
\hat{f}(\mathbf{x}) = \frac{\alpha}{N + \alpha} \mathcal{T}(\mathbf{x} | \boldsymbol{\mu}_0, \boldsymbol{\Lambda}_0, \nu_0) + \sum_{c=1}^{C} \frac{n_c}{N + \alpha} \mathcal{T}(\mathbf{x} | \boldsymbol{\mu}_c, \boldsymbol{\Lambda}_c, \nu_c)
\end{equation}
where all $t$-distributions are obtained by analytically integrating out the Gaussian mean and precision parameters. Specifically, $(\boldsymbol{\mu}_c,\boldsymbol{\Lambda}_c,\nu_c)$ are derived from the updated posterior NW distribution of an existing cluster $c$, while $(\boldsymbol{\mu}_0, \boldsymbol{\Lambda}_0, \nu_0)$ are derived from the NW prior baseline ($\mathbf{G}_0$).

However, greedy sequential assignment is sensitive to the ordering of the observations. Reducing this order dependence would ideally require evaluating different permutations of the data, whose number grows factorially, making exhaustive evaluation computationally infeasible for large datasets. A practical alternative is to evaluate only a limited number of permutations. Although this substantially reduces computational cost, it may also limit the ability to identify a partition with optimal performance. Thus, the number of permutations introduces a trade-off between computational efficiency and density estimation performance.

Therefore, in this work, we consider a middle ground between computational tractability and the ability to account for posterior uncertainty, namely, VI. Specifically, we employ coordinate ascent variational inference (CAVI) \cite{jordan_blei} and its memoized extension \cite{hughes2013memoized}. VI formulates posterior inference as an optimization problem, seeking a tractable distribution $q$ that approximates the intractable posterior. The problem is solved by minimizing the Kullback-Leibler divergence between the true posterior and the approximation function $q$. CAVI assumes a mean-field factorization of $q$, enabling efficient coordinate-wise optimization. Its application to the stick-breaking representation requires a finite truncation level $T$ for the approximation function $q$, which provides a finite approximation to the infinite mixture. A sufficiently large $T$ allows the effect of this truncation on the inferred model to become negligible.

Memoized CAVI \cite{hughes2013memoized} extends CAVI to improve computational efficiency for large datasets by processing data in batches and reusing previously computed quantities, thus reducing the number of successive variational updates. It also allows the truncation level $T$ to be adaptively increased or decreased through heuristic procedures. Overall, these approaches provide a practical balance between computational efficiency and posterior uncertainty, avoiding both the computational cost of MCMC and the single-partition approximation inherent to hard-clustering methods. Other VI approaches (see Section \ref{sec:dpm}) are not considered because they either require extensive hyperparameter tuning, which becomes impractical in BNs due to the need to tailor the inference procedure to the separate CPD associated with each node (e.g. SVI), have been shown to underperform CAVI (e.g. collapsed VI \cite{wang2011fast}), or are specifically designed for online data (e.g. \cite{yang2019memorized}).

By restricting the Dirichlet process mixture to a conjugate exponential-family model, namely a multivariate Gaussian likelihood with a NW prior, the mean-field variational updates admit analytical expressions. Although this imposes a parametric form on the individual mixture components, the mixture representation retains substantial flexibility in the resulting predictive density, since multiple components can combine to represent complex, multimodal, and non-Gaussian distributions. Furthermore, a Gamma prior can be placed on the DP concentration parameter $\alpha$, allowing the data to inform the rate at which new clusters are created. Crucially, this conjugate formulation ensures that the approximate posterior predictive density for a new observation $\mathbf{x}$ admits a closed-form expression as a finite mixture of multivariate Student's $t$-distributions:

\begin{equation}
\hat{f}(\mathbf{x}) \approx \sum_{t=1}^{T} \mathbb{E}_q[\pi_t] \mathcal{T}(\mathbf{x} \mid \boldsymbol{\mu}_t, \boldsymbol{\Lambda}_t, \nu_t),
\end{equation}
where $T$ is the stick-breaking truncation level, $\mathbb{E}_q[\pi_t]$ are the expected variational mixture weights, and $(\boldsymbol{\mu}_t, \boldsymbol{\Lambda}_t, \nu_t)$ are the parameters of the Student's predictive distribution induced by the variational NW posterior for component $t$, obtained by integrating out the Gaussian mean and precision parameters. See the original works of \cite{jordan_blei,hughes2013memoized} for more details.

Given the posterior inference procedure described above, the second issue concerns its extension to CPD modeling. For this purpose, we follow the formulation of SPBN CKDE (see Definition \ref{definition3:1} and Equation \ref{eq:CKDE}). Due to the Gaussian kernel assumption and the relationship between the joint and marginal bandwidths established in Equation \ref{eq:bandwidth_relationship}, the CKDE formulation is equivalent to first estimating the JPD of the child and its parents, $(X_j,\mathbf{X}_{Pa(j)})$, using a multivariate KDE and then conditioning this joint density on the parent variables \cite{atienza2022semiparametric}. This equivalence motivates the same construction for the proposed DPM-based CPD. Instead of modeling the conditional distribution directly, we model the JPD of the child and its parents and subsequently obtain the CPD by conditioning on the parents.

Under the conjugate Gaussian formulation introduced above, the joint DPM density has a closed-form finite-mixture representation in terms of multivariate Student's $t$-distributions. Since the conditional distribution of a multivariate Student's $t$-distribution is also available in closed form, the CPD can be obtained analytically by conditioning the resulting joint mixture on the parent variables, without requiring numerical integration.

This approach is also supported by \cite{quintana2022dependent}, which introduces the dependent Dirichlet process to model a variable conditional on exogenous variables. The authors distinguish this setting from the endogenous case, for which they advocate modeling the JPD of the variables and subsequently obtaining the conditional distribution. In the context of BNs, CPD modeling corresponds to the latter setting. Therefore, we estimate the JPD of \((X_j,\mathbf{X}_{Pa(j)})\) using the DPM and obtain the CPD of \(X_j\) by conditioning on \(\mathbf{X}_{Pa(j)}\). 

The resulting finite mixture representation provides a tractable formulation for deriving the conditional distribution. Combined with the variational inference procedures described above, this yields a computationally efficient non-parametric CPD estimator that preserves the probabilistic structure of the DPM. Given a child variable $X_j$ and its parent variables $\mathbf{X}_{Pa(j)}$, the proposed DPM-based conditional estimator is formally defined in Definition \ref{definition:VDPM_conditional}. 

\begin{df}
\label{definition:VDPM_conditional}
Let $X_j$ be a random variable and $\mathbf{X}_{Pa(j)}$ be its associated vector of parent variables of dimension $|Pa(j)|$. Let $q$ be the approximate posterior distribution obtained via VI for a DPM model with a NW prior given $\mathcal{D}$ a realized dataset of $N$ i.i.d. observations. The variational DPM estimator for the conditional density of $X_j$ given $\mathbf{X}_{Pa(j)}$ is defined as:
\begin{equation}
\hat{f}(x_j\mid \mathbf{x}_{Pa(j)}) = \frac{\sum_{t=1}^{T} \mathbb{E}_q[\pi_t] \mathcal{T}((x_j, \mathbf{x}_{Pa(j)}) \mid \boldsymbol{\mu}_t, \boldsymbol{\Lambda}_t, \nu_t)}{\sum_{t=1}^{T} \mathbb{E}_q[\pi_t] \mathcal{T}(\mathbf{x}_{Pa(j)} \mid \boldsymbol{\mu}_{Pa,t}, \boldsymbol{\Lambda}_{Pa,t}, \nu_t)},
\label{eq:VDPM_conditional}
\end{equation}
where $T$ represents the number of mixture components utilized by the variational approximation, and $\mathcal{T}(\cdot \mid \boldsymbol{\mu}, \boldsymbol{\Lambda}, \nu)$ is the multivariate Student's $t$-density obtained by marginalizing the component-specific Gaussian parameters against the variational posterior $q$. The term $\mathbb{E}_q[\pi_t]$ denotes the expected mixture weight of the $t$-th component. The joint location vector $\boldsymbol{\mu}_t$ and scale matrix $\boldsymbol{\Lambda}_t$ for each component partition into the marginal parameters for the parents as follows:
\begin{equation}
\begin{aligned}
\boldsymbol{\mu}_t &= \begin{bmatrix} \mu_{j,t} \\ \boldsymbol{\mu}_{Pa,t} \end{bmatrix}, \quad \text{where} \quad \mu_{j,t}\in\mathbb{R}, \hspace{1.5mm} \boldsymbol{\mu}_{Pa,t}\in \mathbb{R}^{|Pa(j)|} \\
\boldsymbol{\Lambda}_t &= \begin{bmatrix}
\Lambda_{j,t} & \boldsymbol{\Lambda}_{j,Pa,t}^{\mathsf{T}} \\
\boldsymbol{\Lambda}_{j,Pa,t} & \boldsymbol{\Lambda}_{Pa,t}
\end{bmatrix}, \quad \text{where} \quad \Lambda_{j,t}\in\mathbb{R}, \hspace{1.5mm} \boldsymbol{\Lambda}_{j, Pa,t}\in \mathbb{R}^{|Pa(j)|}, \hspace{1.5mm} \boldsymbol{\Lambda}_{Pa,t}\in\mathbb{M}^{|Pa(j)|\times |Pa(j)|}.
\end{aligned}
\label{eq:VDPM_partition}
\end{equation}

\end{df}

With this definition, the theoretical framework of the proposed DPM-SPBN is established, incorporating both parametric and non-parametric approaches for CPD learning. 

\subsection{Structure learning}
\label{sec:structure_learning_theory}
In this section, we address the learning of DAG structures, completing the formulation of the proposed DPM-SPBN framework. For structure learning, we retain the framework proposed for SPBNs in \cite{atienza2022semiparametric}. The original work adapts both the HC and PC algorithms to the semiparametric setting. We retain these adaptations for DPM-SPBNs and replace the CKDE formulation with the proposed DPM-based CPD of Definition \ref{definition:VDPM_conditional}.

In the HC algorithm, the original framework introduces a \texttt{TYPE-CHANGE} operator in addition to the standard addition, removal, and reversal operators. This operator allows the CPD type of a node to switch between the parametric LG and non-parametric formulations during structure learning. For DPM-SPBNs, we retain this operator and replace the CKDE with the DPM-based CPD.

Standard scores such as BIC and AIC are not directly applicable because the number of free parameters of non-parametric CPDs is not well defined. Consequently, the corresponding complexity penalty cannot be computed. Moreover, training-data evaluation can cause overfitting for KDE-based CPDs, as each observation contributes a kernel centered on itself. The DPM formulation avoids this specific issue, but the number of effective parameters remains difficult to define. We therefore retain the $k$-fold cross-validated log-likelihood (LL) score proposed in the original SPBN framework to guide the search for the network structure. The complete dataset is first partitioned into a training set $\mathcal{D}^{train}$ and a hold-out validation set $\mathcal{D}^{val}$. At each step of the structural search, candidate moves are evaluated using a $k$-fold cross-validation score strictly within $\mathcal{D}^{train}$:
\begin{equation}
\mathcal{S}_{CV}^{k}(\mathcal{D}^{train},\mathcal{G})=\sum_{m=1}^{k}\mathcal{L}(\mathcal{G},\theta_{\mathcal{I}_{train}^{m}}:\mathcal{D}_{\mathcal{I}_{test}^{m}}),
\end{equation}
where $\mathcal{L}(\cdot)$ denotes the LL function, and $\theta_{\mathcal{I}_{train}^{m}}$ represents the parameters estimated on the $m$-th training fold and evaluated on the corresponding internal test fold $\mathcal{D}_{\mathcal{I}_{test}^{m}}$. 

Once the best candidate move is selected, the parameters $\theta_{\mathcal{D}^{train}}$ for the newly accepted graph $\mathcal{G}$ are re-estimated using the entirety of $\mathcal{D}^{train}$. To prevent the sequential search from overfitting, an early-stopping criterion then evaluates this updated model on the separate validation set:
\begin{equation}
\mathcal{S}_{validation}(\mathcal{D}^{train},\mathcal{D}^{val},\mathcal{G})=\mathcal{L}(\mathcal{G},\theta_{\mathcal{D}^{train}}:\mathcal{D}^{val}).
\end{equation}

If this validation score degrades, the structural search sequence is halted to preserve the model's generalization ability.

For the constraint based approach, the original SPBN framework adapts the MPC version of the PC stable algorithm \cite{colombo2014order}, which provides order independent structure learning. It uses non parametric conditional independence tests, comparing CMIknn and RCoT, with RCoT preferred due to its lower computational cost. Finally, the greedy HC algorithm with only the \texttt{TYPE-CHANGE} operator is applied to select the CPD type for each node. We retain this procedure for DPM-SPBNs.

\section{Experiments}
\label{sec:4}
In this section, we evaluate and compare the proposed DPM-SPBN with the original KDE-based SPBN and state-of-the-art BNs for modeling continuous variables through experiments on parameter estimation and structure learning. To assess the performance, we conducted experiments on both synthetic and real-world datasets. Synthetic datasets allow us to control the data generating process and evaluate performance against known underlying distributions, while real-world datasets from the UCI Machine Learning Repository\footnote{https://archive.ics.uci.edu/} provide a benchmark for assessing the practical applicability of BNs.

\begin{figure}[htbp]
    \centering
    \includegraphics[width=0.8\textwidth]{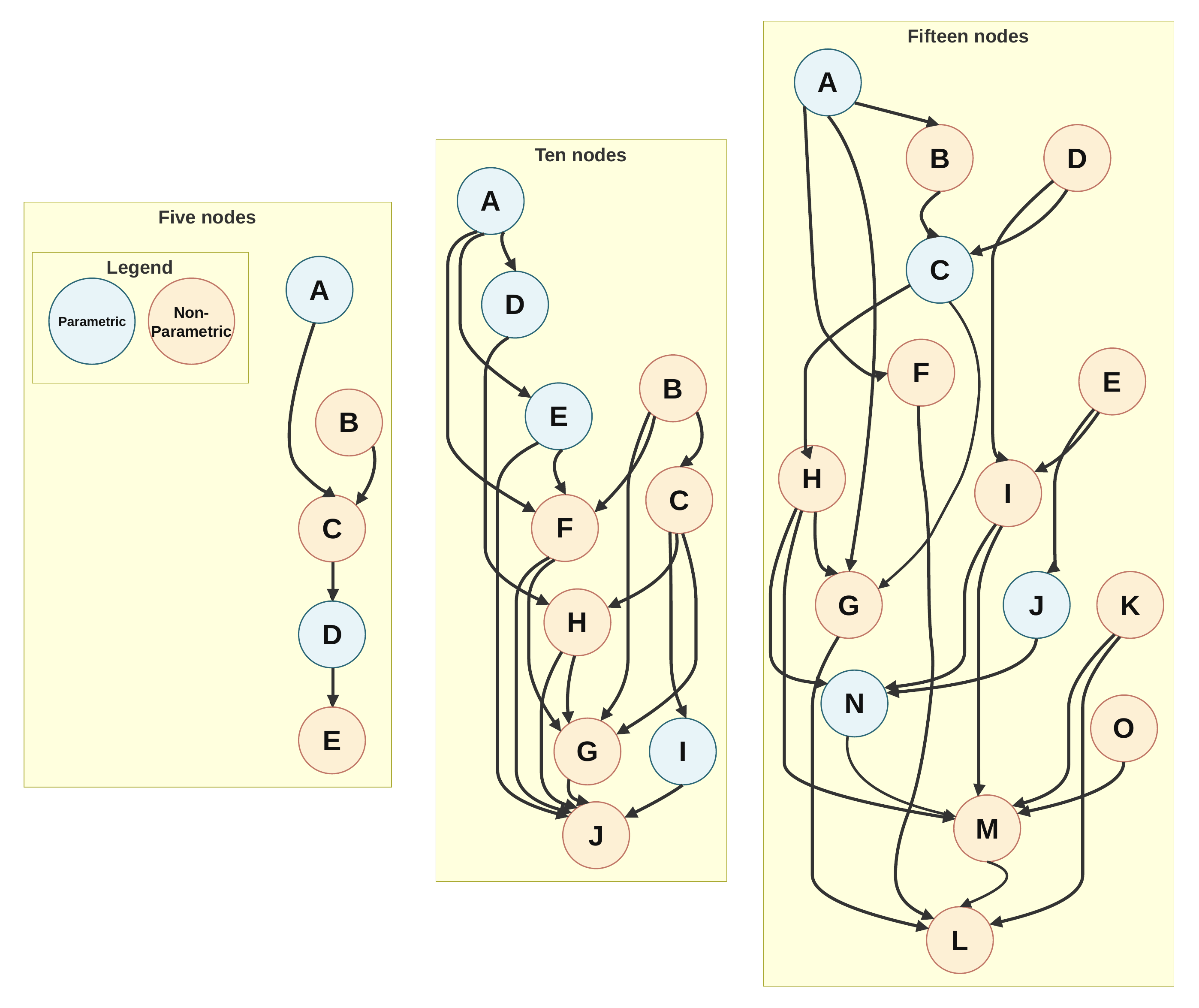}
    \caption{Bayesian networks defined for the experimental setup of the work with sample sizes of 5, 10 and 15 nodes.}
    \label{fig:experiments_networks}
\end{figure}

For synthetic experiments, we consider three network structures consisting of $5$, $10$, and $15$ nodes, following the experimental design in \cite{alejandre2026bandwidth} (see Figure \ref{fig:experiments_networks}). For each network structure, we consider multiple sample sizes and underlying distributions with different levels of smoothness, thereby covering a range of data generating conditions (see Table \ref{tab:data_generation} for an example).

\begin{table}[h]
\centering
\renewcommand{\arraystretch}{1.25}
\setlength{\tabcolsep}{8pt}
\footnotesize
\begin{tabular}{@{}p{0.15\linewidth} p{0.83\linewidth}@{}}
\toprule
\centering \textbf{Smooth} &
\[
\begin{aligned}
    &\bullet \; f(x_1) = \mathcal{N}(0, 1), \\
    &\bullet \; f(x_2) = 0.5 \mathcal{N}(-2, 2^2) + 0.5 \mathcal{N}(2, 2^2), \\
    &\bullet \; f(x_3 \mid x_1, x_2) = \mathcal{N}(x_1 x_2, 1), \\
    &\bullet \; f(x_4 \mid x_3) = \mathcal{N}(10 + 0.8 x_3, 0.5^2), \\
    &\bullet \; f(x_5 \mid x_4) = \mathcal{N}\!\left(\frac{1}{1 + e^{-x_4}}, 0.5^2\right)
\end{aligned}
\] \\

\midrule

\centering \textbf{Medium} &
\[
\begin{aligned}
    &\bullet \; f(x_1) = \mathcal{N}(0, 3^2), \\
    &\bullet \; f(x_2) = 0.25 \mathcal{N}(-8, 1^2) + 0.25 \mathcal{N}(8, 1^2) + 0.25 \mathcal{N}(-4, 0.5^2) + 0.25 \mathcal{N}(4, 0.5^2),\\
    &\bullet \; f(x_3 \mid x_1, x_2) =
    w_1 \mathcal{N}(-0.5 x_1, 1) +
    w_2 \mathcal{N}(0.5 x_2, 1) +
    w_3 \mathcal{N}(0, 1),\\
    &\hspace{1cm}
    w_1 = \frac{x_1^2}{x_1^2 + x_2^2 + 2|x_1 x_2|}, \quad
    w_2 = \frac{x_2^2}{x_1^2 + x_2^2 + 2|x_1 x_2|}, \quad
    w_3 = \frac{2|x_1 x_2|}{x_1^2 + x_2^2 + 2|x_1 x_2|},\\
    &\bullet \; f(x_4 \mid x_3) = \mathcal{N}(0.5 x_3, 0.5^2), \\
    &\bullet \; f(x_5 \mid x_4) = w \mathcal{N}(-2, 0.5^2) + (1 - w)\mathcal{N}(2, 0.5^2), \quad
    w = \frac{1}{1 + \exp(-x_4 / 3)}
\end{aligned}
\] \\

\midrule

\centering \textbf{Rough} &
\[
\begin{aligned}
    &\bullet \; f(x_1) = \mathcal{N}(0, 3^2), \\
    &\bullet \; f(x_2) = 0.2 \mathcal{N}(-8, 1^2) + 0.2 \mathcal{N}(8, 1^2) + 0.2 \mathcal{N}(-4, 0.5^2) + 0.2 \mathcal{N}(0, 0.5^2) + 0.2 \mathcal{N}(4, 0.5^2),\\
    &\bullet \; f(x_3 \mid x_1, x_2) =
    w_1 \mathcal{N}(-0.5 x_1^2, 1) +
    w_2 \mathcal{N}(0.5 x_2^2, 1) +
    w_3 \mathcal{N}(0.5 x_1, 1) +
    w_4 \mathcal{N}(-0.5 x_2, 1),\\
    &\hspace{1cm}
    w_1 = \frac{x_1^2}{x_1^2 + x_2^2 + |x_1| + |x_2|}, \quad
    w_2 = \frac{x_2^2}{x_1^2 + x_2^2 + |x_1| + |x_2|},\\
    &\hspace{1cm}
    w_3 = \frac{|x_1|}{x_1^2 + x_2^2 + |x_1| + |x_2|}, \quad
    w_4 = \frac{|x_2|}{x_1^2 + x_2^2 + |x_1| + |x_2|},\\
    &\bullet \; f(x_4 \mid x_3) = \mathcal{N}(0.5 x_3, 0.5^2), \\
    &\bullet \; f(x_5 \mid x_4) =
    w_1 \mathcal{N}(-2, 0.5^2) +
    w_2 \mathcal{N}(0, 0.5^2) +
    w_3 \mathcal{N}(2, 0.5^2),\\
    &\hspace{1cm}
    w_1 = \frac{1}{(1 + \exp(-x_4 / 3)) * 2}, \quad
    w_2 = \frac{1}{(1 + \exp(-x_4 / 3)) * 2}, \quad
    w_3 = 1 - w_1 - w_2
\end{aligned}
\] \\

\midrule

\centering \textbf{Nonparametric} &
\[
\begin{aligned}
    &\bullet \; f(x_1) = 0.25 \mathcal{N}(-4, 0.5^2) + 0.25 \mathcal{N}(-2, 0.5^2) + 0.25 \mathcal{N}(2, 0.5^2) + 0.25 \mathcal{N}(4, 0.5^2), \\
    &\bullet \; f(x_2) = 0.5 \mathcal{N}(-2, 2^2) + 0.5 \mathcal{N}(2, 2^2), \\
    &\bullet \; f(x_3 \mid x_1, x_2) = \mathcal{N}(x_1 x_2, 1), \\
    &\bullet \; f(x_4 \mid x_3) = 0.5 \mathcal{N}(10 + 0.8 x_3, 0.5^2) + 0.5 \mathcal{N}(10 - 0.8 x_3, 0.5^2), \\
    &\bullet \; f(x_5 \mid x_4) = \mathcal{N}\!\left(\frac{1}{1 + e^{-x_4}}, 0.5^2\right)
\end{aligned}
\] \\

\bottomrule
\end{tabular}
\caption{Example of data generation mechanisms for the \textbf{Smooth}, \textbf{Medium}, \textbf{Rough}, and \textbf{Nonparametric} densities in the 5 node network.}
\label{tab:data_generation}
\end{table}

Since the proposed method introduces a new approach to CPD modeling, we first evaluate it within the parameter learning paradigm using known and fixed network structures. These experiments compare DPM-SPBN with state-of-the-art BNs to model continuous distributions. We also compare DPM-SPBN with standalone DPM models to assess the benefits of combining the BN framework with DPM-based inference over only DPM density estimation.

Furthermore, CPD modeling can directly affect structure learning in SPBNs, both through score computation in HC and through the selection of CPD types (Gaussian or non-parametric) in the PC algorithm. Therefore, we conduct additional experiments to investigate how the proposed CPD modeling approach affects structure learning and to quantify the impact of CPD modeling on the overall SPBN learning process.

For the experimental evaluation, we extended the Python package \texttt{PyBNesian} \cite{atienza2022pybnesian}, which provides support for SPBNs, to incorporate the proposed DPM-SPBN model. The extension integrates DPM-SPBN throughout the package, enabling parameter and structure learning, density evaluation, and sampling from the JPD encoded by the BN within a unified framework. All code is available in Github repository\footnote{https://github.com/Victor-Alejandre/DPM-SPBN.git}.

For the statistical comparison of the methods, we follow the framework proposed by \cite{benavoli2016should}. This framework argues against classical rank-based procedures, whose results can depend substantially on the number of algorithms being compared \cite{garcia2008extension}. Instead, we perform pairwise comparisons and account for multiple testing using the Bergmann-Hommel correction to control the family-wise error rate. For pairwise comparisons, we use a paired median permutation test, applying the dependent-sample framework described by \cite{bonnini2024review}. This test provides a non-parametric assessment of performance differences without requiring assumptions about the underlying error distribution.

\subsection{Parameter learning experiments}
\label{sec:cpd_experiments}
We first evaluate CPD estimation under a fixed network structure $\mathcal{G}$. DPM-SPBN is compared with several non-parametric approaches for continuous BNs, including SPBNs using NR and UCV bandwidth selection \cite{alejandre2026bandwidth}, SPBNs based on semiparametric linear regression (LRSPBN) \cite{Boukabour17122021}, pair-copula BNs with Gaussian copulas (GPCBNs) \cite{hanea2015non}, and Gaussian copula BNs (GCBNs) \cite{elidan2010copula} (see Section \ref{sec:bayesian_networks} and Figure \ref{fig:continous CPD}). We exclude KDE-based networks \cite{hofmann1995discovering} due to their close methodological relationship with SPBNs, and Gaussian process networks \cite{Friedman2000GaussianPN} due to their prohibitive computational cost in the experimental setting considered. Similarly, we restrict the comparison to Gaussian copula models among copula-based approaches for computational and theoretical reasons. Infinite-mixture-based BNs \cite{ickstadt2010nonparametric} are also excluded because their structure learning formulation does not obtain a standard BN representation with explicit CPDs. 
 
For each of the three network structures ($5$, $10$, and $15$ nodes), we define a fixed topology with varying numbers of parent nodes and a combination of Gaussian and non-parametric CPDs, resulting in heterogeneous CPD configurations. For the mixed configurations, conditional densities are specified at three levels of smoothness: smooth, medium, and rough. To accommodate methods that cannot handle mixed CPDs, we also consider a fully non-parametric (NP) configuration for each topology, preserving the same structure and assigning non-parametric CPDs to all nodes.

For each experimental configuration, we consider five sample sizes, $N \in \{200, 2000, 10000, 50000, 100000\}$, covering a broad range of practical scenarios, with a total of $60$ experimental scenarios. For each combination of network structure, CPD configuration, density setting, and sample size, we generate $50$ independent training-test dataset pairs, with $10000$ observations in each test set to provide a large-sample approximation to the population for metric evaluation. All experiments are fully reproducible; the complete implementation and experimental configurations are available in the accompanying GitHub repository.

We use LL as the primary performance metric. In the BN framework, where models define probability densities, LL directly measures the density assigned to the observations and, therefore, provides a natural measure of overall density estimation performance across the test dataset. Moreover, because the true data-generating distribution is known in the synthetic experiments, we assessed density estimation accuracy using the absolute LL error relative to the true distribution. We then compare training times and evaluate the trade-off between absolute LL error and computational cost. Prediction times are also studied, given that one of the main advantages of DPM-SPBNs is the potential to achieve low inference times by using a reduced number of mixture components.

To ensure a fair and reproducible comparison, hyperparameters were set according to standardized defaults and the methodological properties of each model. For LRSPBN, the statistical linearity test  used $\alpha=0.05$ to partition parent sets into linear and non-parametric components. For GPCBNs, the parent nodes were ordered alphabetically; given the known DAG structure. This fixed ordering provides a consistent parameterization without affecting the underlying JPD. GCBNs do not need any parameter configuration.

For DPMs, we used CAVI, CAVI with Gamma prior on concentration parameter (GPCAVI) and Memoized. CAVI and GPCAVI have a truncation level of $T=20$ components. For CAVI, the concentration parameter is initialized at $1/T$, providing a weak regularization against over-clustering while allowing the component weights to be primarily determined by the observed data. For GPCAVI and Memoized, the concentration parameter is instead assigned a $\Gamma(1,1)$ hyperprior, allowing it to be informed by the data during inference. In Memoized, this facilitates the heuristic adaptation of the truncation level through the merge and split operations.

For CAVI and GPCAVI, we use a convergence tolerance of $0.001$ and a maximum of $1000$ coordinate ascent iterations. Memoized uses mini-batches of size $1000$, a maximum of $100$ iterations, and a birth proposal frequency of $5$, following the computational design of the underlying methodology.

The NW prior is initialized using the empirical mean and covariance of each dataset to define its location and inverse scale parameters, respectively. The mean precision is fixed at $1$, corresponding to the influence of a single pseudo-observation, while the degrees of freedom are set to the number of features, satisfying the minimum condition for a proper, non-singular distribution. These choices provide a weakly informative initialization while limiting prior influence on the variational updates.

\textbf{Error analysis}. Table \ref{tab:results_fixed_structure} summarizes the results for all cases considered. The table compares the three proposed methods for DPM posterior inference: CAVI, GPCAVI, and Memoized. For each case, we select the best-performing baseline (last column) among UCV, NR, GPCBN, GCBN, and LRSPBN based on the median error performance. For each method, we report the median and the $25$th and $75$th percentiles (forming the interquartile interval (IQI)) computed across the $50$ independent training-test dataset replicas. The method with the best median performance per experiment case is highlighted in bold.

Table \ref{tab:results_fixed_structure} shows that the proposed DPM-SPBN methods consistently achieve better median performance than the best alternative method. At least one of the three proposed inference methods obtains the best median performance in all cases except for the setting ($1$ out of $60$) with $200$ observations, medium density, and $5$-node network. More remarkably, in all but two cases, the IQIs of all three proposed DPM-SPBN inference methods are completely separated from the IQI of the best alternative method, with no overlap between the corresponding intervals. Thus, the observed performance of the proposed methods is not only better in terms of the median, but their central $50\%$ of results lies entirely outside the corresponding range of the best competing alternative. This substantial separation is observed consistently across the experimental settings and highlights the magnitude and consistency of the performance differences between DPM-SPBN and the competing approaches.

\begin{table}[htbp]
\centering
\caption{Synthetic data experiments Absolute error summary for the LL across different node structures, densities and sample sizes in parameter learning experiments. Each case displays median and [IQI] for each method with the best median highlighted in bold. Best alternative method for each case is display between parenthesis.}
\label{tab:results_fixed_structure}
\tiny
\renewcommand{\arraystretch}{0.9}
\setlength{\tabcolsep}{3pt}
\begin{tabular}{@{}c p{1cm} l r@{.}l r@{.}l r@{.}l r@{.}l@{}}
\toprule
\textbf{Nodes} & \textbf{Density} & \textbf{\makecell{Sample \\ Size}} & \multicolumn{2}{c}{\textbf{\makecell{SPBN-CAVI \\ IQI}}} & \multicolumn{2}{c}{\textbf{\makecell{SPBN-GPCAVI \\ IQI}}} & \multicolumn{2}{c}{\textbf{\makecell{SPBN-MEMOIZED \\ IQI}}} & \multicolumn{2}{c}{\textbf{\makecell{Best \\ Alternative \\ IQI}}} \\
\midrule
\multirow{20}{*}{5} & \multirow{5}{2.2cm}{Smooth} & 200 & $\mathbf{3548}$ & $\mathbf{44}~[3388.59, 3740.85]$ & $3647$ & $55~[3416.09, 3888.17]$ & $4285$ & $07~[3862.44, 4566.23]$ & $6085$ & $68~[5910.16, 6344.54]$ (NR) \\
 &  & 2000 & $\mathbf{896}$ & $\mathbf{37}~[847.70, 939.15]$ & $950$ & $71~[902.42, 1007.55]$ & $1047$ & $26~[984.11, 1147.39]$ & $2550$ & $46~[2160.48, 2941.39]$ (UCV) \\
 &  & 10000 & $383$ & $94~[363.52, 409.47]$ & $351$ & $49~[321.34, 372.01]$ & $\mathbf{346}$ & $\mathbf{07}~[325.27, 374.63]$ & $1066$ & $21~[920.25, 1249.90]$ (UCV) \\
 &  & 50000 & $251$ & $95~[234.60, 280.55]$ & $200$ & $69~[185.11, 208.88]$ & $\mathbf{117}$ & $\mathbf{36}~[101.12, 124.94]$ & $434$ & $73~[383.04, 521.34]$ (UCV) \\
 &  & 100000 & $241$ & $00~[223.69, 258.95]$ & $190$ & $76~[171.64, 199.12]$ & $\mathbf{75}$ & $\mathbf{14}~[65.70, 87.74]$ & $472$ & $16~[421.11, 568.19]$ (UCV) \\
\cmidrule(l){2-11}
 & \multirow{5}{2.2cm}{Medium} & 200 & $6431$ & $66~[6238.14, 6623.57]$ & $6558$ & $40~[6260.33, 6882.96]$ & $7276$ & $29~[7101.96, 7376.25]$ & $\mathbf{3622}$ & $\mathbf{71}~[3375.82, 4680.40]$ (UCV) \\
 &  & 2000 & $\mathbf{997}$ & $\mathbf{44}~[961.42, 1060.32]$ & $1071$ & $72~[955.09, 1140.82]$ & $1059$ & $74~[1017.03, 1110.33]$ & $1273$ & $61~[1163.86, 1391.88]$ (UCV) \\
 &  & 10000 & $\mathbf{350}$ & $\mathbf{41}~[336.56, 362.95]$ & $350$ & $91~[325.30, 395.51]$ & $517$ & $32~[477.95, 577.40]$ & $667$ & $14~[630.52, 717.73]$ (UCV) \\
 &  & 50000 & $250$ & $74~[228.91, 264.92]$ & $250$ & $37~[228.35, 275.39]$ & $\mathbf{162}$ & $\mathbf{34}~[152.00, 180.92]$ & $357$ & $42~[325.24, 373.80]$ (UCV) \\
 &  & 100000 & $236$ & $44~[219.59, 250.68]$ & $242$ & $17~[222.97, 257.16]$ & $\mathbf{97}$ & $\mathbf{06}~[78.17, 122.03]$ & $396$ & $63~[377.66, 427.70]$ (UCV) \\
\cmidrule(l){2-11}
 & \multirow{5}{2.2cm}{Rough} & 200 & $\mathbf{1}$ & $\mathbf{95}~[1.92, 1.98] \times 10^{4}$ & $1$ & $97~[1.93, 2.00] \times 10^{4}$ & $2$ & $10~[2.02, 2.18] \times 10^{4}$ & $2$ & $34~[2.31, 2.37] \times 10^{4}$ (NR) \\
 &  & 2000 & $6970$ & $41~[6880.55, 7055.42]$ & $7096$ & $50~[6935.96, 7529.66]$ & $\mathbf{6863}$ & $\mathbf{02}~[6659.19, 7092.33]$ & $1$ & $02~[0.96, 1.08] \times 10^{4}$ (UCV) \\
 &  & 10000 & $3197$ & $07~[3106.08, 3323.87]$ & $3565$ & $97~[3329.07, 4095.03]$ & $\mathbf{3032}$ & $\mathbf{84}~[2931.56, 3155.91]$ & $5395$ & $93~[5108.71, 5948.88]$ (UCV) \\
 &  & 50000 & $1903$ & $30~[1774.78, 2136.54]$ & $2336$ & $13~[2212.98, 2535.85]$ & $\mathbf{1365}$ & $\mathbf{06}~[1325.46, 1438.36]$ & $3081$ & $13~[2955.31, 3198.98]$ (UCV) \\
 &  & 100000 & $1634$ & $91~[1476.37, 1815.43]$ & $1984$ & $32~[1836.30, 2143.91]$ & $\mathbf{867}$ & $\mathbf{68}~[784.29, 1003.26]$ & $3084$ & $24~[2976.83, 3189.34]$ (UCV) \\
\cmidrule(l){2-11}
 & \multirow{5}{2.2cm}{NP} & 200 & $\mathbf{1}$ & $\mathbf{61}~[1.56, 1.64] \times 10^{4}$ & $1$ & $64~[1.60, 1.73] \times 10^{4}$ & $1$ & $85~[1.77, 1.92] \times 10^{4}$ & $2$ & $88~[2.21, 3.74] \times 10^{4}$ (UCV) \\
 &  & 2000 & $\mathbf{3978}$ & $\mathbf{61}~[3859.46, 4090.39]$ & $4468$ & $16~[4291.69, 7172.50]$ & $4448$ & $62~[4318.41, 4544.27]$ & $6791$ & $14~[6260.54, 7587.83]$ (UCV) \\
 &  & 10000 & $1711$ & $95~[1607.64, 1805.66]$ & $2101$ & $96~[2017.19, 2172.24]$ & $\mathbf{1348}$ & $\mathbf{39}~[1300.29, 1392.48]$ & $3227$ & $21~[2935.85, 3520.98]$ (UCV) \\
 &  & 50000 & $839$ & $59~[803.92, 875.97]$ & $820$ & $85~[793.57, 866.49]$ & $\mathbf{445}$ & $\mathbf{14}~[418.88, 463.45]$ & $1503$ & $75~[1453.72, 1592.77]$ (UCV) \\
 &  & 100000 & $696$ & $86~[673.64, 725.97]$ & $688$ & $13~[665.93, 717.78]$ & $\mathbf{273}$ & $\mathbf{14}~[261.08, 295.23]$ & $1701$ & $59~[1678.54, 1737.73]$ (UCV) \\
\midrule
\multirow{20}{*}{10} & \multirow{5}{2.2cm}{Smooth} & 200 & $\mathbf{4}$ & $\mathbf{50}~[4.37, 4.88] \times 10^{4}$ & $4$ & $57~[4.40, 5.00] \times 10^{4}$ & $4$ & $88~[4.75, 5.22] \times 10^{4}$ & $8$ & $90~[8.80, 9.09] \times 10^{4}$ (GPCBN) \\
 &  & 2000 & $\mathbf{2}$ & $\mathbf{58}~[2.38, 2.80] \times 10^{4}$ & $2$ & $71~[2.46, 3.04] \times 10^{4}$ & $2$ & $64~[2.46, 2.91] \times 10^{4}$ & $8$ & $52~[8.38, 8.74] \times 10^{4}$ (GPCBN) \\
 &  & 10000 & $1$ & $83~[1.68, 2.10] \times 10^{4}$ & $1$ & $92~[1.77, 2.22] \times 10^{4}$ & $\mathbf{1}$ & $\mathbf{72}~[1.59, 2.05] \times 10^{4}$ & $8$ & $22~[6.86, 14.77] \times 10^{4}$ (NR) \\
 &  & 50000 & $1$ & $41~[1.31, 1.60] \times 10^{4}$ & $1$ & $42~[1.31, 1.59] \times 10^{4}$ & $\mathbf{1}$ & $\mathbf{20}~[1.08, 1.39] \times 10^{4}$ & $6$ & $36~[6.01, 7.20] \times 10^{4}$ (NR) \\
 &  & 100000 & $1$ & $29~[1.20, 1.40] \times 10^{4}$ & $1$ & $27~[1.17, 1.38] \times 10^{4}$ & $\mathbf{1}$ & $\mathbf{02}~[0.91, 1.15] \times 10^{4}$ & $6$ & $05~[5.72, 6.52] \times 10^{4}$ (NR) \\
\cmidrule(l){2-11}
 & \multirow{5}{2.2cm}{Medium} & 200 & $\mathbf{1}$ & $\mathbf{74}~[1.68, 1.79] \times 10^{4}$ & $1$ & $75~[1.69, 1.83] \times 10^{4}$ & $1$ & $75~[1.68, 1.84] \times 10^{4}$ & $2$ & $12~[2.09, 2.15] \times 10^{4}$ (NR) \\
 &  & 2000 & $\mathbf{6805}$ & $\mathbf{54}~[6710.30, 6962.91]$ & $7216$ & $55~[7054.00, 7358.69]$ & $7003$ & $82~[6811.87, 7212.95]$ & $1$ & $16~[1.12, 1.20] \times 10^{4}$ (UCV) \\
 &  & 10000 & $4404$ & $52~[4299.29, 4495.83]$ & $4529$ & $91~[4433.71, 4626.42]$ & $\mathbf{3773}$ & $\mathbf{76}~[3649.23, 3879.84]$ & $6543$ & $53~[6320.94, 6781.90]$ (UCV) \\
 &  & 50000 & $3733$ & $74~[3623.95, 3899.85]$ & $3777$ & $65~[3648.96, 3916.55]$ & $\mathbf{1887}$ & $\mathbf{94}~[1838.79, 1964.36]$ & $4015$ & $46~[3907.11, 4179.17]$ (UCV) \\
 &  & 100000 & $3684$ & $19~[3559.91, 3808.12]$ & $3693$ & $32~[3550.30, 3809.31]$ & $\mathbf{1520}$ & $\mathbf{47}~[1414.80, 1630.74]$ & $4944$ & $53~[4806.78, 5142.05]$ (UCV) \\
\cmidrule(l){2-11}
 & \multirow{5}{2.2cm}{Rough} & 200 & $1$ & $00~[0.95, 1.05] \times 10^{5}$ & $1$ & $01~[0.95, 1.06] \times 10^{5}$ & $\mathbf{9}$ & $\mathbf{95}~[9.50, 10.54] \times 10^{4}$ & $1$ & $42~[1.38, 1.49] \times 10^{5}$ (GPCBN) \\
 &  & 2000 & $\mathbf{7}$ & $\mathbf{65}~[7.43, 7.80] \times 10^{4}$ & $7$ & $77~[7.46, 7.88] \times 10^{4}$ & $7$ & $83~[7.61, 7.96] \times 10^{4}$ & $1$ & $29~[1.27, 1.32] \times 10^{5}$ (NR) \\
 &  & 10000 & $\mathbf{6}$ & $\mathbf{40}~[6.36, 6.45] \times 10^{4}$ & $6$ & $44~[6.38, 6.48] \times 10^{4}$ & $6$ & $51~[6.43, 6.58] \times 10^{4}$ & $1$ & $18~[1.17, 1.19] \times 10^{5}$ (NR) \\
 &  & 50000 & $5$ & $51~[5.42, 5.56] \times 10^{4}$ & $5$ & $50~[5.43, 5.55] \times 10^{4}$ & $\mathbf{5}$ & $\mathbf{24}~[5.21, 5.28] \times 10^{4}$ & $1$ & $04~[0.90, 1.60] \times 10^{5}$ (UCV) \\
 &  & 100000 & $5$ & $12~[5.06, 5.20] \times 10^{4}$ & $5$ & $12~[5.06, 5.20] \times 10^{4}$ & $\mathbf{4}$ & $\mathbf{76}~[4.71, 4.79] \times 10^{4}$ & $7$ & $44~[6.94, 8.73] \times 10^{4}$ (UCV) \\
\cmidrule(l){2-11}
 & \multirow{5}{2.2cm}{NP} & 200 & $\mathbf{1}$ & $\mathbf{87}~[1.82, 1.96] \times 10^{5}$ & $1$ & $88~[1.82, 1.98] \times 10^{5}$ & $1$ & $90~[1.85, 1.99] \times 10^{5}$ & $2$ & $52~[2.48, 2.61] \times 10^{5}$ (GCBN) \\
 &  & 2000 & $\mathbf{1}$ & $\mathbf{50}~[1.45, 1.54] \times 10^{5}$ & $1$ & $51~[1.46, 1.55] \times 10^{5}$ & $1$ & $52~[1.47, 1.55] \times 10^{5}$ & $2$ & $50~[2.45, 2.52] \times 10^{5}$ (GPCBN) \\
 &  & 10000 & $1$ & $30~[1.27, 1.33] \times 10^{5}$ & $1$ & $30~[1.27, 1.33] \times 10^{5}$ & $\mathbf{1}$ & $\mathbf{28}~[1.25, 1.31] \times 10^{5}$ & $2$ & $16~[2.08, 2.40] \times 10^{5}$ (NR) \\
 &  & 50000 & $1$ & $17~[1.16, 1.21] \times 10^{5}$ & $1$ & $17~[1.15, 1.21] \times 10^{5}$ & $\mathbf{1}$ & $\mathbf{11}~[1.09, 1.14] \times 10^{5}$ & $2$ & $00~[1.95, 2.07] \times 10^{5}$ (NR) \\
 &  & 100000 & $1$ & $13~[1.11, 1.16] \times 10^{5}$ & $1$ & $13~[1.11, 1.16] \times 10^{5}$ & $\mathbf{1}$ & $\mathbf{03}~[1.01, 1.06] \times 10^{5}$ & $1$ & $73~[1.35, 8.13] \times 10^{5}$ (UCV) \\
\midrule
\multirow{20}{*}{15} & \multirow{5}{2.2cm}{Smooth} & 200 & $2$ & $06~[1.99, 2.10] \times 10^{4}$ & $2$ & $16~[2.08, 2.22] \times 10^{4}$ & $\mathbf{1}$ & $\mathbf{64}~[1.60, 1.70] \times 10^{4}$ & $3$ & $14~[3.10, 3.20] \times 10^{4}$ (NR) \\
 &  & 2000 & $6746$ & $93~[6576.09, 6911.38]$ & $6978$ & $88~[6789.74, 7207.55]$ & $\mathbf{6378}$ & $\mathbf{66}~[6205.62, 6732.35]$ & $8394$ & $20~[8019.20, 8808.78]$ (UCV) \\
 &  & 10000 & $3032$ & $25~[2973.89, 3082.45]$ & $3090$ & $74~[3026.23, 3158.60]$ & $\mathbf{2943}$ & $\mathbf{10}~[2863.75, 3028.14]$ & $5070$ & $53~[4911.53, 5288.00]$ (UCV) \\
 &  & 50000 & $1714$ & $61~[1670.46, 1770.64]$ & $1732$ & $50~[1678.65, 1781.11]$ & $\mathbf{1197}$ & $\mathbf{54}~[1168.42, 1259.74]$ & $3273$ & $45~[3127.79, 3470.15]$ (UCV) \\
 &  & 100000 & $1554$ & $53~[1522.04, 1614.42]$ & $1560$ & $84~[1524.65, 1608.65]$ & $\mathbf{832}$ & $\mathbf{15}~[809.54, 869.27]$ & $5415$ & $72~[5298.88, 5591.00]$ (UCV) \\
\cmidrule(l){2-11}
 & \multirow{5}{2.2cm}{Medium} & 200 & $\mathbf{1}$ & $\mathbf{77}~[1.73, 1.84] \times 10^{5}$ & $1$ & $78~[1.76, 1.85] \times 10^{5}$ & $1$ & $79~[1.74, 1.86] \times 10^{5}$ & $2$ & $57~[2.51, 2.65] \times 10^{5}$ (GPCBN) \\
 &  & 2000 & $\mathbf{1}$ & $\mathbf{42}~[1.39, 1.45] \times 10^{5}$ & $1$ & $44~[1.40, 1.47] \times 10^{5}$ & $1$ & $43~[1.40, 1.46] \times 10^{5}$ & $2$ & $45~[2.42, 2.50] \times 10^{5}$ (GPCBN) \\
 &  & 10000 & $\mathbf{1}$ & $\mathbf{27}~[1.24, 1.30] \times 10^{5}$ & $1$ & $29~[1.25, 1.32] \times 10^{5}$ & $1$ & $28~[1.25, 1.32] \times 10^{5}$ & $2$ & $32~[2.09, 2.80] \times 10^{5}$ (NR) \\
 &  & 50000 & $1$ & $17~[1.15, 1.19] \times 10^{5}$ & $1$ & $17~[1.15, 1.19] \times 10^{5}$ & $\mathbf{1}$ & $\mathbf{15}~[1.14, 1.18] \times 10^{5}$ & $1$ & $98~[1.93, 2.14] \times 10^{5}$ (NR) \\
 &  & 100000 & $1$ & $14~[1.12, 1.16] \times 10^{5}$ & $1$ & $14~[1.12, 1.15] \times 10^{5}$ & $\mathbf{1}$ & $\mathbf{11}~[1.09, 1.13] \times 10^{5}$ & $1$ & $94~[1.90, 2.00] \times 10^{5}$ (NR) \\
\cmidrule(l){2-11}
 & \multirow{5}{2.2cm}{Rough} & 200 & $\mathbf{2}$ & $\mathbf{67}~[2.64, 2.69] \times 10^{5}$ & $2$ & $67~[2.64, 2.69] \times 10^{5}$ & $2$ & $70~[2.67, 2.74] \times 10^{5}$ & $3$ & $20~[3.17, 3.30] \times 10^{5}$ (NR) \\
 &  & 2000 & $2$ & $24~[2.23, 2.25] \times 10^{5}$ & $2$ & $26~[2.25, 2.28] \times 10^{5}$ & $\mathbf{2}$ & $\mathbf{24}~[2.23, 2.25] \times 10^{5}$ & $2$ & $88~[2.87, 2.89] \times 10^{5}$ (NR) \\
 &  & 10000 & $2$ & $07~[2.06, 2.07] \times 10^{5}$ & $2$ & $07~[2.06, 2.08] \times 10^{5}$ & $\mathbf{2}$ & $\mathbf{05}~[2.05, 2.06] \times 10^{5}$ & $2$ & $74~[2.73, 2.74] \times 10^{5}$ (NR) \\
 &  & 50000 & $1$ & $92~[1.91, 1.93] \times 10^{5}$ & $1$ & $92~[1.91, 1.93] \times 10^{5}$ & $\mathbf{1}$ & $\mathbf{87}~[1.86, 1.88] \times 10^{5}$ & $2$ & $62~[2.62, 2.62] \times 10^{5}$ (NR) \\
 &  & 100000 & $1$ & $86~[1.86, 1.87] \times 10^{5}$ & $1$ & $86~[1.85, 1.87] \times 10^{5}$ & $\mathbf{1}$ & $\mathbf{80}~[1.78, 1.80] \times 10^{5}$ & $2$ & $41~[2.13, 3.73] \times 10^{5}$ (UCV) \\
\cmidrule(l){2-11}
 & \multirow{5}{2.2cm}{NP} & 200 & $2$ & $15~[2.09, 2.25] \times 10^{4}$ & $2$ & $27~[2.18, 2.35] \times 10^{4}$ & $\mathbf{1}$ & $\mathbf{67}~[1.62, 1.77] \times 10^{4}$ & $3$ & $11~[3.05, 3.18] \times 10^{4}$ (GPCBN) \\
 &  & 2000 & $5531$ & $21~[5394.22, 5699.32]$ & $6061$ & $30~[5794.30, 7164.76]$ & $\mathbf{4787}$ & $\mathbf{47}~[4691.37, 4884.18]$ & $8453$ & $53~[8051.78, 8853.65]$ (UCV) \\
 &  & 10000 & $2175$ & $23~[2122.07, 2216.47]$ & $2659$ & $04~[2575.88, 2712.18]$ & $\mathbf{1861}$ & $\mathbf{86}~[1794.66, 1953.14]$ & $4965$ & $96~[4802.70, 5162.69]$ (UCV) \\
 &  & 50000 & $976$ & $53~[931.06, 1011.22]$ & $1094$ & $24~[1048.98, 1139.44]$ & $\mathbf{833}$ & $\mathbf{01}~[796.32, 873.50]$ & $3081$ & $68~[2963.18, 3163.68]$ (UCV) \\
 &  & 100000 & $771$ & $57~[740.16, 824.45]$ & $876$ & $61~[841.28, 930.74]$ & $\mathbf{546}$ & $\mathbf{36}~[523.64, 576.63]$ & $8161$ & $79~[7924.55, 8428.63]$ (UCV) \\
\bottomrule
\end{tabular}
\end{table}

Among the proposed inference methods, CAVI generally achieves the best performance for smaller sample sizes, whereas Memoized tends to outperform the other two methods as the sample size increases. GPCAVI generally exhibits the weakest performance among the three proposed inference methods, as it does not achieve the best median performance in any of the scenarios considered.

Among the alternative methods, SPBN with UCV bandwidth selection is generally the best-performing approach, followed by SPBN with the less complex NR bandwidth selector. GPCBN and GCBN achieve the best median performance among the alternatives in a few scenarios with small sample sizes.

\begin{figure}[htbp]
    \centering
    \includegraphics[width=0.8\textwidth]{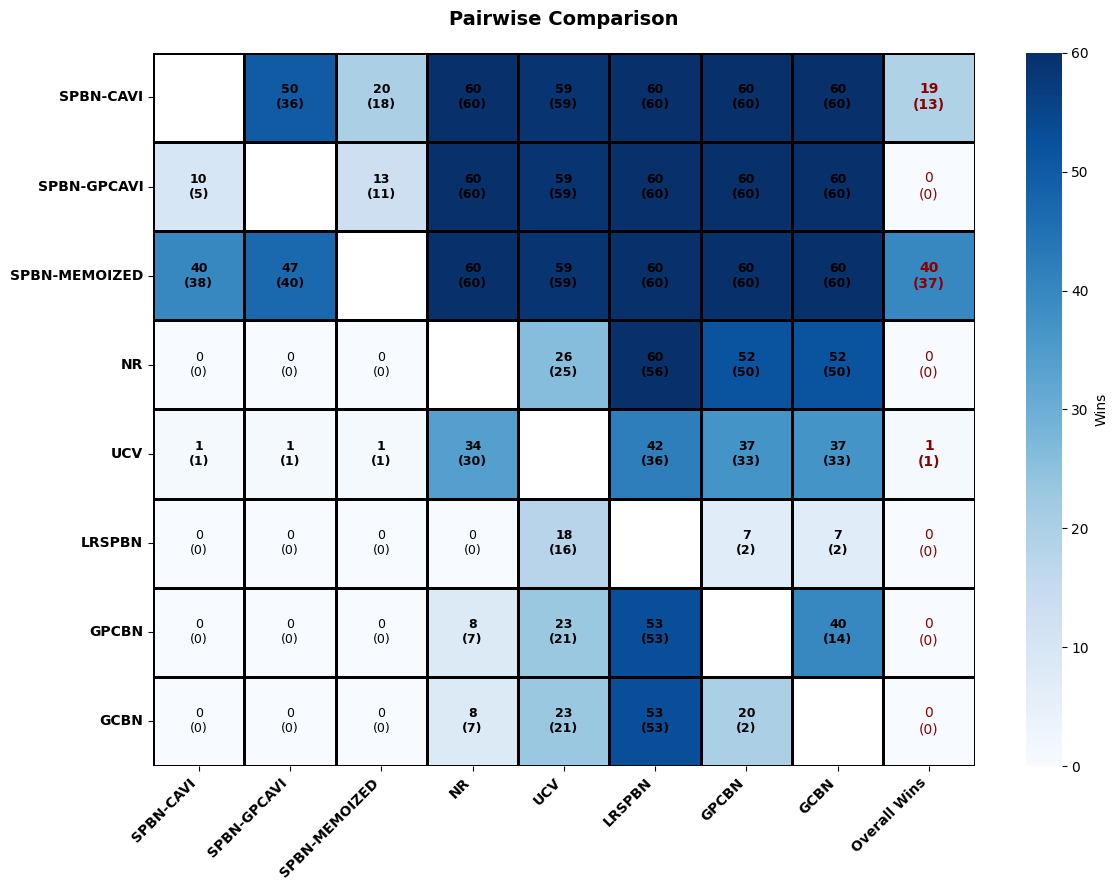}
    \caption{Pairwise comparison of state-of-the-art BNs for continous variables modeling and DPM-SPBN methods in parameter learning experiments. Each cell shows the number of cases in which the row method achieves a better median performance than the column method, with statistically significant differences reported in parentheses. Last column summarizes the overall and significant wins.}
    \label{fig:fixed_structure_comparison}
\end{figure}

\textbf{Statistical analysis}. Table \ref{tab:statistical_summary} summarizes the statistical comparisons performed to assess the significance of the performance differences observed in Table \ref{tab:results_fixed_structure}. All eight methods considered in the experiments were compared pairwise, with multiple-testing correction applied to the complete set of comparisons. For each method, the table first reports the number of cases in which it achieves the lowest median absolute error across all methods (i.e., the overall wins), followed by a disaggregation of these wins according to the statistically significant pairwise comparisons with the other methods. For clarity, only statistically significant comparisons involving the proposed inference methods and the best-performing alternative in each case are reported. Figure \ref{fig:fixed_structure_comparison} provides a complete disaggregated summary of the pairwise comparisons.

\begin{table}[htbp]
\centering
\caption{Statistical significance of the proposed methods when achieving the lowest median absolute error. For each method, the number of cases in which it achieves the lowest median absolute error across all methods (total wins) and, among these, the number of wins with statistically significant pairwise differences are reported.}
\label{tab:statistical_summary}
\scriptsize
\begin{tabular}{@{}l c c c c c@{}}
\toprule
\textbf{Proposed Method} & \textbf{Total Wins} & \textbf{Sig. vs CAVI} & \textbf{Sig. vs GPCAVI} & \textbf{Sig. vs MEMOIZED} & \textbf{Sig. vs Best Alt.} \\
\midrule
SPBN-CAVI & 19 & - & 14 & 17 & 19 \\
SPBN-GPCAVI & 0 & 0 & - & 0 & 0 \\
SPBN-MEMOIZED & 40 & 38 & 39 & - & 40 \\
\bottomrule
\end{tabular}
\end{table}

The proposed methods show a statistically significant difference in median performance from the best alternative in all cases in which a proposed method achieves the best median performance, corresponding to $59$ out of the $60$ experimental scenarios. Among the proposed inference methods, CAVI and Memoized are frequently significantly different from the other proposed methods when they achieve the best median performance. In contrast, GPCAVI does not achieve the best median performance in any of the scenarios considered. Together with the performance patterns observed in Table \ref{tab:results_fixed_structure}, these results suggest that CAVI is generally the preferable inference method for smaller sample sizes, while Memoized is more advantageous as the sample size increases.

\begin{figure}[htbp]
    \centering
    \includegraphics[width=1.03\textwidth]{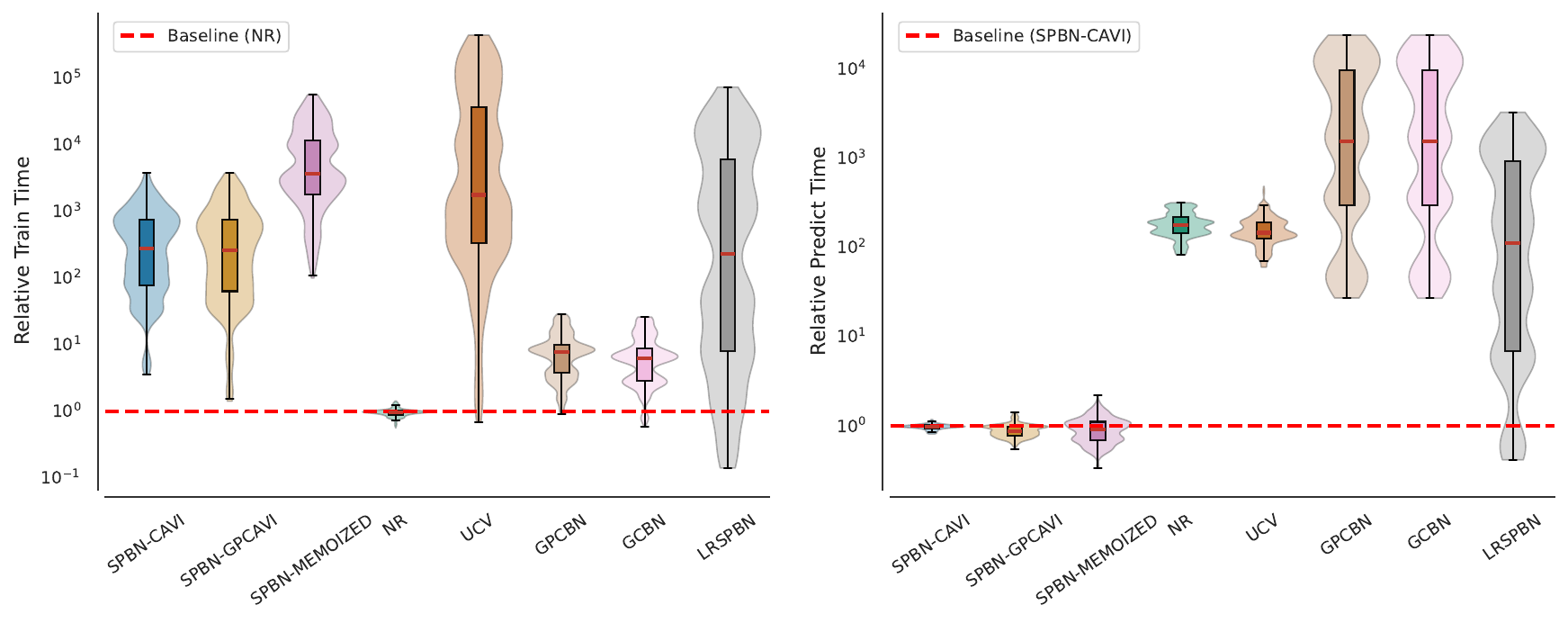}
    \caption{Parameter learning experiments distribution of relative computation times across network structures and sample sizes, with times averaged over density levels. The reference value corresponds to NR for training and SPBN-CAVI for prediction. Relative times are shown on a logarithmic scale.}
    \label{fig:computation_runtime}
\end{figure}

\textbf{Computational analysis}. Table \ref{tab:computation_time_full} and Figure \ref{fig:computation_runtime} summarize the computational times, averaging over network densities for each combination of network size and sample size. Table \ref{tab:computation_time_full} reports in detail the resulting training and prediction times, with the minimum time for each case shown in bold. Figure \ref{fig:computation_runtime} instead summarizes the relative times across all network- and sample-size combinations, using baseline NR for training and SPBN-CAVI for prediction. Among the proposed DPM inference methods, CAVI, GPCAVI, and Memoized have comparable prediction times. Compared to alternative BN methods, DPM-SPBN reduces inference times by approximately $10^2$–$10^3$ times, substantially lowering prediction cost.

\begin{sidewaystable}[htbp]
\centering
\caption{Parameter learning experiments average training and prediction computation times (s) across network structures and sample sizes, averaged over all density levels. Minimum times for each network-size and sample-size combination are shown in bold.}
\label{tab:computation_time_full}
\footnotesize
\setlength{\tabcolsep}{3pt}
\begin{tabular}{@{}c l r@{.}l r@{.}l r@{.}l r@{.}l r@{.}l r@{.}l r@{.}l r@{.}l r@{.}l r@{.}l@{}}
\toprule
\multirow{3}{*}{\textbf{Nodes}} & \multirow{3}{*}{\textbf{Method}} & \multicolumn{20}{c}{\textbf{Sample Size}} \\
\cmidrule(l){3-22}
& & \multicolumn{4}{c}{\textbf{200}} & \multicolumn{4}{c}{\textbf{2000}} & \multicolumn{4}{c}{\textbf{10000}} & \multicolumn{4}{c}{\textbf{50000}} & \multicolumn{4}{c}{\textbf{100000}} \\
\cmidrule(lr){3-6} \cmidrule(lr){7-10} \cmidrule(lr){11-14} \cmidrule(lr){15-18} \cmidrule(l){19-22}
& & \multicolumn{2}{c}{\textbf{Train (s)}} & \multicolumn{2}{c}{\textbf{Predict (s)}} & \multicolumn{2}{c}{\textbf{Train (s)}} & \multicolumn{2}{c}{\textbf{Predict (s)}} & \multicolumn{2}{c}{\textbf{Train (s)}} & \multicolumn{2}{c}{\textbf{Predict (s)}} & \multicolumn{2}{c}{\textbf{Train (s)}} & \multicolumn{2}{c}{\textbf{Predict (s)}} & \multicolumn{2}{c}{\textbf{Train (s)}} & \multicolumn{2}{c}{\textbf{Predict (s)}} \\
\midrule
\multirow{8}{*}{5} 
    & SPBN-CAVI      & 0 & 021 & 0 & 0050 & 0 & 219 & 0 & 0044 & 0 & 897 & 0 & 0047 & 3 & 67 & 0 & 0047 & 7 & 34 & 0 & 0055 \\
    & SPBN-GPCAVI    & 0 & 030 & \textbf{0} & \textbf{0038} & 0 & 185 & 0 & 0037 & 0 & 997 & 0 & 0040 & 3 & 02 & 0 & 0041 & 6 & 26 & 0 & 0046 \\
    & SPBN-MEMOIZED  & 1 & 50 & 0 & 0039 & 2 & 13 & \textbf{0} & \textbf{0030} & 9 & 11 & \textbf{0} & \textbf{0036} & 68 & 8 & \textbf{0} & \textbf{0038} & 135 & 0 & \textbf{0} & \textbf{0044} \\
    & NR             & 0 & 010 & 0 & 501 & \textbf{0} & \textbf{0023} & 0 & 500 & \textbf{0} & \textbf{0040} & 0 & 786 & \textbf{0} & \textbf{018} & 0 & 728 & \textbf{0} & \textbf{030} & 0 & 893 \\
    & UCV            & 0 & 031 & 0 & 481 & 0 & 264 & 0 & 473 & 1 & 81 & 0 & 651 & 18 & 2 & 0 & 629 & 97 & 8 & 5 & 65 \\
    & GPCBN          & 0 & 0061 & 0 & 183 & 0 & 0062 & 1 & 60 & 0 & 015 & 7 & 68 & 0 & 059 & 36 & 8 & 0 & 133 & 75 & 5 \\
    & GCBN           & 0 & 0042 & 0 & 182 & 0 & 0058 & 1 & 60 & 0 & 014 & 7 & 68 & 0 & 057 & 36 & 8 & 0 & 131 & 75 & 6 \\
    & LRSPBN           & \textbf{0} & \textbf{0038} & 0 & 021 & 0 & 050 & 0 & 148 & 0 & 994 & 0 & 632 & 28 & 3 & 3 & 50 & 84 & 3 & 4 & 94 \\
\midrule
\multirow{8}{*}{10} 
    & SPBN-CAVI      & 0 & 041 & 0 & 0079 & 0 & 480 & 0 & 011 & 2 & 36 & 0 & 010 & 13 & 8 & 0 & 0087 & 30 & 4 & 0 & 010 \\
    & SPBN-GPCAVI    & 0 & 044 & \textbf{0} & \textbf{0070} & 0 & 391 & 0 & 0091 & 2 & 22 & 0 & 0090 & 15 & 5 & \textbf{0} & \textbf{0078} & 35 & 5 & \textbf{0} & \textbf{0091} \\
    & SPBN-MEMOIZED  & 3 & 04 & 0 & 0087 & 8 & 02 & \textbf{0} & \textbf{0084} & 38 & 7 & \textbf{0} & \textbf{0075} & 278 & 7 & 0 & 012 & 600 & 9 & 0 & 0100 \\
    & NR             & \textbf{0} & \textbf{0012} & 1 & 44 & \textbf{0} & \textbf{0042} & 1 & 45 & \textbf{0} & \textbf{0045} & 2 & 22 & \textbf{0} & \textbf{013} & 2 & 05 & \textbf{0} & \textbf{018} & 2 & 64 \\
    & UCV            & 0 & 430 & 1 & 42 & 6 & 32 & 1 & 42 & 49 & 8 & 2 & 08 & 965 & 3 & 1 & 96 & 3900 & 00 & 17 & 2 \\
    & GPCBN          & 0 & 011 & 0 & 387 & 0 & 015 & 3 & 43 & 0 & 038 & 16 & 5 & 0 & 151 & 78 & 5 & 0 & 334 & 162 & 3 \\
    & GCBN           & 0 & 0076 & 0 & 388 & 0 & 012 & 3 & 43 & 0 & 033 & 16 & 5 & 0 & 142 & 78 & 5 & 0 & 321 & 162 & 3 \\
    & LRSPBN           & 0 & 010 & 0 & 051 & 0 & 249 & 0 & 355 & 5 & 28 & 1 & 85 & 142 & 4 & 10 & 6 & 421 & 3 & 15 & 6 \\
\midrule
\multirow{8}{*}{15} 
    & SPBN-CAVI      & 0 & 070 & 0 & 010 & 0 & 846 & 0 & 013 & 3 & 00 & 0 & 012 & 16 & 2 & 0 & 016 & 28 & 4 & 0 & 012 \\
    & SPBN-GPCAVI    & 0 & 083 & \textbf{0} & \textbf{010} & 0 & 976 & 0 & 013 & 3 & 35 & 0 & 012 & 17 & 6 & \textbf{0} & \textbf{016} & 32 & 9 & 0 & 012 \\
    & SPBN-MEMOIZED  & 4 & 29 & 0 & 013 & 7 & 80 & \textbf{0} & \textbf{011} & 39 & 3 & \textbf{0} & \textbf{011} & 256 & 8 & 0 & 018 & 601 & 1 & \textbf{0} & \textbf{012} \\
    & NR             & \textbf{0} & \textbf{0050} & 1 & 92 & \textbf{0} & \textbf{0060} & 1 & 94 & \textbf{0} & \textbf{0073} & 2 & 04 & \textbf{0} & \textbf{022} & 2 & 65 & \textbf{0} & \textbf{021} & 3 & 36 \\
    & UCV            & 0 & 500 & 1 & 92 & 6 & 54 & 1 & 94 & 58 & 4 & 2 & 04 & 1100 & 00 & 2 & 58 & 4700 & 00 & 20 & 4 \\
    & GPCBN          & 0 & 016 & 0 & 564 & 0 & 022 & 5 & 06 & 0 & 053 & 24 & 3 & 0 & 175 & 107 & 9 & 0 & 500 & 242 & 9 \\
    & GCBN           & 0 & 011 & 0 & 566 & 0 & 017 & 5 & 06 & 0 & 046 & 24 & 3 & 0 & 157 & 107 & 9 & 0 & 482 & 243 & 0 \\
    & LRSPBN           & 0 & 013 & 0 & 066 & 0 & 309 & 0 & 469 & 7 & 04 & 2 & 58 & 263 & 2 & 17 & 5 & 631 & 2 & 20 & 9 \\
\bottomrule
\end{tabular}
\end{sidewaystable}

This reduction in inference time comes at the cost of higher training times for the proposed methods, although these remain feasible, particularly for CAVI and GPCAVI. For Memoized, contrary to the original claims, the computational cost increases significantly with both the sample size and the number of nodes. This behavior is particularly relevant when considering the use of these methods within structure learning procedures such as HC, where CPDs must be repeatedly fitted throughout the search. Among the alternative methods, SPBN with the NR bandwidth selector achieves the lowest training times, approximately $10^2$-$10^3$ times lower than those of the proposed methods. However, as shown in the previous results, this method does not achieve the best density-estimation accuracy among the alternative approaches and performs worse than the proposed DPM-SPBN methods.

Considering that the training times of the proposed methods remain feasible, while inference times are reduced by approximately two orders of magnitude and the log-density error is significantly lower in almost all cases, the proposed DPM-SPBN methods provide a more suitable approach for CPD learning in BNs. The increase in computational cost observed for Memoized with dataset and network size also anticipates an important consideration for its application within computationally intensive structure learning procedures.

\begin{figure}[htbp]
    \centering
    \includegraphics[width=0.8\textwidth]{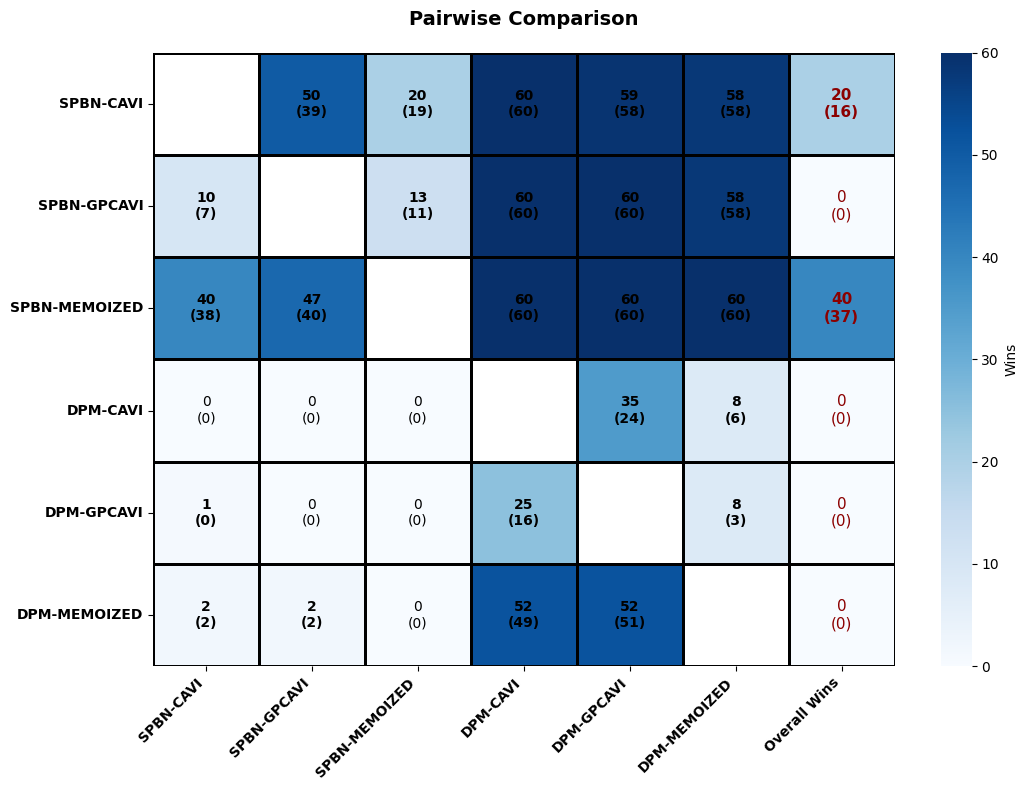}
    \caption{Pairwise comparison of DPM-SPBN and standalone DPM methods in parameter learning experiments. Each cell shows the number of cases in which the row method achieves a better median performance than the column method, with statistically significant differences cases reported in parentheses. Last column summarizes the overall wins and significant wins out of those against all competitors.}
    \label{fig:DPMvsSPBN}
\end{figure}

Finally, Figure \ref{fig:DPMvsSPBN} presents a summary of pairwise performance and statistical comparisons between the proposed DPM-SPBN methods and their corresponding standalone DPM counterparts, using the same inference procedures but without the BN structure. The comparison is performed under the same experimental scenarios as in the previous experiments. For each pair of methods, the figure reports the number of cases in which the row method achieves better median performance than the column method, with the number of statistically significant differences shown in parentheses. The last column summarizes the overall number of wins and statistically significant wins for each method across all pairwise comparisons.

The results clearly demonstrate the advantage of combining BNs with DPM-based inference. In almost all pairwise comparisons, DPM-SPBN outperforms the corresponding standalone DPM methods, regardless of the inference procedure, with statistically significant differences in median performance, as shown by the dark-blue cells on the right side of the matrix. Moreover, the DPM-SPBN methods achieve the best overall performance in all $60$ experimental scenarios, with statistically significant differences in every case. These results highlight the benefit of the BN factorization, which provides a substantial advantage over fitting a single DPM to the full JPD in the considered settings.

\subsection{Structure learning experiments}
\label{sec:structure_experiments}
In this section, we study the effect of the proposed non-parametric CPD model on structure learning within SPBNs. Structure learning is a distinct problem from CPD modeling, but the two can be either closely related or largely independent depending on the structure learning approach. Given the extensive literature on structure learning and the different ways in which CPD models can interact with the structure search, we focus specifically on the SPBN framework, where the CPD model is directly involved in the structure learning procedure. 

Therefore, we compare the proposed DPM-based CPD with existing SPBN approaches based on KDEs and bandwidth selection, using HC with the cross-validation score (see Section~\ref{sec:structure_learning_theory}), where CPDs are estimated throughout the search. GCBN, GPCBN, and LRSPBN are not considered, as they use different CPD formulations and structure-learning settings (see Section \ref{sec:bayesian_networks}). This choice is further supported by the comparatively poor CPD estimation performance of these approaches in the previous section.

For the structure learning experiments, we use the same experimental setting as in the parameter learning experiments (see Section \ref{sec:cpd_experiments}) with respect to network structures and sample sizes. However, we exclude the NP density setting because SPBN is designed to accommodate both parametric and non-parametric CPDs, making the smooth, medium, and rough density settings more appropriate for evaluating structure learning with mixed CPD types. This results in a total of $45$ experimental scenarios.

For each experimental configuration, we define a grid of hyperparameters to reproduce the real approach search usually used for structure learning. We consider $5$ and $10$ folds with a $80/20$ training-validation split for the score, together with patience values of $0$ and $5$. The patience parameter allows the search to continue for a predefined number of iterations despite a deterioration in the score, reducing the likelihood of terminating the search prematurely at a local optimum. The tolerance for the stopping criterion is fixed at $\epsilon=0.01$. For each configuration, we generate an independent training dataset and a test dataset of $10000$ observations. We repeat this sampling procedure $10$ times per configuration. For each method and experimental configuration, the hyperparameter setting is selected according to the best median LL across these $10$ repetitions. Absolute LL error is not used for hyperparameter selection, as the underlying distribution is unknown in practical applications; this criterion therefore reflects the model-selection procedure available in real-world settings.

In this experimental design, we compare SPBNs using the NR and UCV bandwidth selectors with DPM-SPBN using CAVI and GPCAVI with the same settings as in Section \ref{sec:cpd_experiments}. Memoized is excluded because the computational cost of evaluating the proposed hyperparameter grid across all settings would be too high, although its use may remain feasible for individual configurations. For the same computational consideration, UCV is evaluated only for sample sizes of $200$, $2000$, and $10000$ observations, as its computational cost becomes prohibitive for the larger sample sizes considered.

We evaluate the structural accuracy of the learned networks using the structural Hamming distance (SHD) \cite{tsamardinos2006max} with respect to the ground-truth structure, which accounts for the class of equivalent DAGs. Since the underlying data generating distribution is known in our synthetic experiments, we additionally computed the absolute LL error. While SHD directly evaluates structural recovery, the LL error allows us to assess how accurately the learned network approximates the underlying distribution and to examine its relationship with structural accuracy. This is particularly relevant because structure learning based on score optimization and search identifies a structure that maximizes the selected score, but does not necessarily guarantee the most accurate approximation of the underlying distribution.

Table \ref{tab:shd_abserr_summary} summarizes both metrics. For each method, we report the IQI of the SHD values, with the best median SHD highlighted in bold. The LL column indicates whether a method achieves the lowest median absolute LL error for each experimental configuration. The two final rows report the number of wins for each method. The first reports the number of SHD wins out of the total number of cases, excluding the four ties, together with the number of these SHD wins that also achieve the lowest absolute LL error. The second reports the total number of cases with the lowest median LL error out of all experimental cases, independently of the SHD results.

The results show that the estimation performance of the proposed CAVI and GPCAVI methods (Section \ref{sec:cpd_experiments}) translates into improved structure recovery when combined with HC. The proposed methods achieve the best SHD in approximately $75\%$ of the cases. GPCAVI performs slightly worse than CAVI, with fewer SHD wins and fewer cases in which these wins are also accompanied by the best estimation accuracy. Nevertheless, both methods perform well, and no clear pattern determines which method performs better for a given configuration.

NR provides a relevant alternative for structure recovery, achieving the best SHD in most of the remaining cases, particularly for larger networks. However, NR does not achieve the best density estimation accuracy in any of these cases. In contrast, UCV achieves the best density estimation accuracy in a few cases but the best SHD in only one comparison. These results illustrate how structure recovery and density estimation can yield different method rankings, with the best-scoring structure not necessarily being the ground truth and the best density estimator not necessarily achieving the best structural recovery.

Overall, the proposed methods provide the strongest performance for HC structure recovery, while NR can be competitive when structure recovery is considered alone. When density estimation is also considered, the proposed methods perform best in nearly all cases. CAVI is generally slightly better than GPCAVI and more frequently achieves the best performance across both structure recovery and density estimation.

Table \ref{tab:abs_err_summary} reports the absolute LL errors in detail, including results for standalone DPM methods. Despite the error in structure recovery, CAVI and GPCAVI outperform the standalone DPM methods in density estimation accuracy in all cases. This result highlights the benefit of incorporating DPMs within BNs and demonstrates their robustness for density estimation.

\begin{table}[htbp]
\centering
\caption{Structural Hamming distance (SHD) and log-likelihood error (LL Err.) across network sizes, densities, and sample sizes in the structure learning experiments. SHD is reported as median [IQR], while LL Err. indicates whether the method achieves (\checkmark) the lowest absolute log-likelihood error or not ($\times$).}
\label{tab:shd_abserr_summary}
\scriptsize
\renewcommand{\arraystretch}{1.1}
\setlength{\tabcolsep}{3pt}
\begin{tabular}{@{}c c l c c c c c c c c@{}}
\toprule
\multirow{2}{*}{\textbf{Nodes}} & \multirow{2}{*}{\textbf{Density}} & \multirow{2}{*}{\textbf{\makecell{Sample \\ Size}}} & \multicolumn{2}{c}{\textbf{SPBN-CAVI}} & \multicolumn{2}{c}{\textbf{SPBN-GPCAVI}} & \multicolumn{2}{c}{\textbf{NR}} & \multicolumn{2}{c}{\textbf{UCV}} \\
\cmidrule(lr){4-5} \cmidrule(lr){6-7} \cmidrule(lr){8-9} \cmidrule(lr){10-11}
& & & \textbf{SHD IQI} & \textbf{LL Err.} & \textbf{SHD IQI} & \textbf{LL Err.} & \textbf{SHD IQI} & \textbf{LL Err.} & \textbf{SHD IQI} & \textbf{LL Err.} \\
\midrule
\multirow{15}{*}{5} & \multirow{5}{*}{Smooth} & 200 & $1.5~[1.0, 5.8]$ & $\times$ & $\mathbf{1.0}~[1.0, 4.8]$ & \checkmark & $4.5~[1.2, 5.8]$ & $\times$ & $4.5~[4.0, 5.8]$ & $\times$ \\
 &  & 2000 & $5.0~[2.0, 5.0]$ & \checkmark & $\mathbf{1.0}~[1.0, 1.8]$ & $\times$ & $1.5~[1.0, 5.0]$ & $\times$ & $2.5~[1.0, 5.0]$ & $\times$ \\
 &  & 10000 & $3.5~[1.0, 6.0]$ & $\times$ & $\mathbf{1.0}~[1.0, 4.2]$ & \checkmark & $4.0~[1.2, 6.8]$ & $\times$ & $2.5~[1.0, 5.0]$ & $\times$ \\
 &  & 50000 & $\mathbf{4.0}~[3.2, 6.0]$ & $\times$ & $\mathbf{4.0}~[3.2, 6.0]$ & \checkmark & $5.0~[3.5, 5.8]$ & $\times$ &  -  &  -  \\
 &  & 100000 & $\mathbf{2.0}~[2.0, 2.8]$ & $\times$ & $3.0~[3.0, 6.0]$ & \checkmark & $5.5~[2.8, 6.0]$ & $\times$ &  -  &  -  \\
\cmidrule(l){2-11}
 & \multirow{5}{*}{Medium} & 200 & $\mathbf{4.0}~[4.0, 4.0]$ & $\times$ & $\mathbf{4.0}~[4.0, 4.0]$ & $\times$ & $\mathbf{4.0}~[4.0, 4.8]$ & $\times$ & $\mathbf{4.0}~[4.0, 5.5]$ & \checkmark \\
 &  & 2000 & $3.5~[2.0, 5.8]$ & \checkmark & $3.0~[0.2, 5.0]$ & $\times$ & $\mathbf{1.5}~[1.0, 2.0]$ & $\times$ & $5.0~[5.0, 5.8]$ & $\times$ \\
 &  & 10000 & $5.0~[5.0, 5.0]$ & \checkmark & $5.0~[5.0, 5.0]$ & $\times$ & $\mathbf{2.0}~[2.0, 2.0]$ & $\times$ & $3.5~[3.0, 4.8]$ & $\times$ \\
 &  & 50000 & $5.5~[5.0, 7.5]$ & \checkmark & $6.0~[5.2, 6.0]$ & $\times$ & $\mathbf{0.0}~[0.0, 0.0]$ & $\times$ &  -  &  -  \\
 &  & 100000 & $6.0~[5.0, 7.0]$ & \checkmark & $6.0~[5.0, 6.0]$ & $\times$ & $\mathbf{0.0}~[0.0, 0.0]$ & $\times$ &  -  &  -  \\
\cmidrule(l){2-11}
 & \multirow{5}{*}{Rough} & 200 & $5.5~[4.0, 7.8]$ & \checkmark & $5.5~[4.0, 6.8]$ & $\times$ & $5.0~[4.0, 5.8]$ & $\times$ & $\mathbf{4.0}~[4.0, 4.8]$ & $\times$ \\
 &  & 2000 & $\mathbf{0.0}~[0.0, 1.0]$ & \checkmark & $2.0~[0.2, 5.8]$ & $\times$ & $4.0~[4.0, 4.8]$ & $\times$ & $5.0~[5.0, 5.8]$ & $\times$ \\
 &  & 10000 & $2.0~[0.0, 4.5]$ & \checkmark & $\mathbf{0.5}~[0.0, 4.0]$ & $\times$ & $4.0~[4.0, 4.0]$ & $\times$ & $4.0~[4.0, 4.0]$ & $\times$ \\
 &  & 50000 & $\mathbf{0.5}~[0.0, 4.8]$ & \checkmark & $2.5~[0.0, 6.8]$ & $\times$ & $4.0~[4.0, 4.0]$ & $\times$ &  -  &  -  \\
 &  & 100000 & $5.0~[1.0, 6.0]$ & \checkmark & $5.0~[1.0, 6.0]$ & $\times$ & $\mathbf{4.0}~[4.0, 4.0]$ & $\times$ &  -  &  -  \\
\midrule
\multirow{15}{*}{10} & \multirow{5}{*}{Smooth} & 200 & $14.5~[10.2, 20.5]$ & \checkmark & $\mathbf{11.5}~[10.0, 14.5]$ & $\times$ & $17.5~[17.0, 18.0]$ & $\times$ & $17.0~[16.2, 18.8]$ & $\times$ \\
 &  & 2000 & $\mathbf{9.0}~[7.2, 10.0]$ & \checkmark & $10.5~[9.2, 11.8]$ & $\times$ & $15.0~[14.0, 16.0]$ & $\times$ & $16.5~[13.2, 18.5]$ & $\times$ \\
 &  & 10000 & $\mathbf{8.5}~[6.2, 10.0]$ & \checkmark & $9.5~[6.2, 11.8]$ & $\times$ & $17.5~[17.0, 18.8]$ & $\times$ & $15.0~[12.5, 17.0]$ & $\times$ \\
 &  & 50000 & $12.0~[9.5, 13.0]$ & $\times$ & $\mathbf{9.0}~[8.0, 10.0]$ & \checkmark & $17.0~[14.0, 17.8]$ & $\times$ &  -  &  -  \\
 &  & 100000 & $\mathbf{10.0}~[9.2, 12.0]$ & $\times$ & $12.5~[11.2, 14.0]$ & \checkmark & $16.0~[14.2, 18.5]$ & $\times$ &  -  &  -  \\
\cmidrule(l){2-11}
 & \multirow{5}{*}{Medium} & 200 & $\mathbf{17.0}~[16.2, 18.8]$ & $\times$ & $18.0~[16.0, 19.8]$ & $\times$ & $\mathbf{17.0}~[16.0, 18.0]$ & $\times$ & $18.0~[16.0, 18.0]$ & \checkmark \\
 &  & 2000 & $\mathbf{16.0}~[13.8, 17.8]$ & \checkmark & $17.0~[15.5, 17.8]$ & $\times$ & $20.5~[18.2, 22.8]$ & $\times$ & $19.0~[18.0, 22.8]$ & $\times$ \\
 &  & 10000 & $\mathbf{12.5}~[12.0, 15.5]$ & \checkmark & $15.0~[14.2, 15.8]$ & $\times$ & $23.0~[22.0, 23.8]$ & $\times$ & $22.0~[20.5, 22.0]$ & $\times$ \\
 &  & 50000 & $19.0~[16.5, 21.2]$ & \checkmark & $\mathbf{17.0}~[14.2, 18.0]$ & $\times$ & $25.0~[24.2, 26.0]$ & $\times$ &  -  &  -  \\
 &  & 100000 & $\mathbf{13.0}~[11.2, 17.8]$ & $\times$ & $16.5~[16.0, 17.8]$ & \checkmark & $26.5~[25.2, 27.0]$ & $\times$ &  -  &  -  \\
\cmidrule(l){2-11}
 & \multirow{5}{*}{Rough} & 200 & $\mathbf{14.0}~[12.2, 15.0]$ & \checkmark & $14.5~[13.0, 17.5]$ & $\times$ & $17.5~[17.0, 19.0]$ & $\times$ & $17.5~[17.0, 19.8]$ & $\times$ \\
 &  & 2000 & $\mathbf{13.0}~[10.2, 15.0]$ & \checkmark & $13.5~[10.0, 15.8]$ & $\times$ & $15.0~[12.2, 18.8]$ & $\times$ & $16.0~[14.5, 18.0]$ & $\times$ \\
 &  & 10000 & $\mathbf{10.0}~[9.2, 10.8]$ & \checkmark & $11.0~[9.0, 14.5]$ & $\times$ & $14.0~[11.0, 15.8]$ & $\times$ & $13.0~[12.2, 18.8]$ & $\times$ \\
 &  & 50000 & $9.5~[8.2, 11.0]$ & $\times$ & $\mathbf{9.0}~[8.2, 13.0]$ & \checkmark & $17.5~[15.2, 20.0]$ & $\times$ &  -  &  -  \\
 &  & 100000 & $\mathbf{9.5}~[7.5, 16.5]$ & $\times$ & $11.0~[8.5, 12.8]$ & \checkmark & $15.5~[13.5, 16.8]$ & $\times$ &  -  &  -  \\
\midrule
\multirow{15}{*}{15} & \multirow{5}{*}{Smooth} & 200 & $19.0~[18.2, 19.8]$ & $\times$ & $18.0~[18.0, 19.8]$ & $\times$ & $\mathbf{16.5}~[16.0, 18.8]$ & $\times$ & $18.0~[17.0, 19.0]$ & \checkmark \\
 &  & 2000 & $19.0~[18.2, 22.0]$ & $\times$ & $\mathbf{15.5}~[13.2, 17.5]$ & $\times$ & $17.0~[15.2, 19.5]$ & $\times$ & $19.0~[18.0, 20.8]$ & \checkmark \\
 &  & 10000 & $\mathbf{15.5}~[15.0, 16.0]$ & \checkmark & $\mathbf{15.5}~[11.2, 17.0]$ & $\times$ & $18.5~[14.2, 19.0]$ & $\times$ & $16.5~[15.2, 17.0]$ & $\times$ \\
 &  & 50000 & $29.0~[21.8, 32.8]$ & \checkmark & $30.5~[26.0, 33.8]$ & $\times$ & $\mathbf{16.5}~[13.0, 19.8]$ & $\times$ &  -  &  -  \\
 &  & 100000 & $25.0~[18.0, 33.0]$ & \checkmark & $21.0~[18.0, 24.0]$ & $\times$ & $\mathbf{20.0}~[19.0, 23.0]$ & $\times$ &  -  &  -  \\
\cmidrule(l){2-11}
 & \multirow{5}{*}{Medium} & 200 & $18.5~[18.0, 22.8]$ & \checkmark & $\mathbf{17.5}~[16.2, 18.8]$ & $\times$ & $22.0~[20.2, 23.0]$ & $\times$ & $23.5~[23.0, 29.5]$ & $\times$ \\
 &  & 2000 & $\mathbf{13.0}~[12.0, 16.2]$ & \checkmark & $14.0~[11.5, 15.8]$ & $\times$ & $20.0~[16.0, 22.5]$ & $\times$ & $27.5~[25.0, 32.0]$ & $\times$ \\
 &  & 10000 & $\mathbf{15.0}~[15.0, 16.5]$ & \checkmark & $15.5~[14.2, 16.0]$ & $\times$ & $20.0~[16.5, 23.5]$ & $\times$ & $22.0~[20.5, 35.0]$ & $\times$ \\
 &  & 50000 & $\mathbf{13.5}~[12.2, 15.8]$ & $\times$ & $14.5~[13.2, 16.0]$ & \checkmark & $14.0~[14.0, 14.8]$ & $\times$ &  -  &  -  \\
 &  & 100000 & $14.5~[13.8, 16.0]$ & $\times$ & $15.0~[12.0, 17.2]$ & \checkmark & $\mathbf{14.0}~[13.8, 20.0]$ & $\times$ &  -  &  -  \\
\cmidrule(l){2-11}
 & \multirow{5}{*}{Rough} & 200 & $19.5~[17.2, 22.5]$ & \checkmark & $\mathbf{18.5}~[15.0, 21.5]$ & $\times$ & $21.5~[18.0, 22.8]$ & $\times$ & $22.5~[19.0, 25.0]$ & $\times$ \\
 &  & 2000 & $\mathbf{13.5}~[10.2, 15.8]$ & \checkmark & $17.0~[15.2, 18.5]$ & $\times$ & $18.0~[8.0, 18.0]$ & $\times$ & $20.0~[14.8, 24.5]$ & $\times$ \\
 &  & 10000 & $20.5~[16.5, 24.8]$ & $\times$ & $22.5~[20.2, 25.2]$ & \checkmark & $\mathbf{9.0}~[8.0, 13.0]$ & $\times$ & $21.0~[17.2, 28.5]$ & $\times$ \\
 &  & 50000 & $16.0~[13.2, 18.8]$ & \checkmark & $16.5~[13.0, 19.0]$ & $\times$ & $\mathbf{8.0}~[8.0, 16.0]$ & $\times$ &  -  &  -  \\
 &  & 100000 & $17.0~[13.8, 21.2]$ & \checkmark & $17.5~[13.5, 23.0]$ & $\times$ & $\mathbf{12.5}~[9.8, 21.0]$ & $\times$ &  -  &  -  \\
\midrule
\multicolumn{3}{c}{\textbf{\makecell{SHD Wins/Total \\ \& \\ Both Wins/SHD Wins}}} & 17/41 & 12/17 & 11/41 & 4/11 & 12/41 & 0/12 & 1/24 & 0/1 \\
\multicolumn{3}{c}{\textbf{LL Wins/Total}} & - & 29/45 & - & 12/45 & - & 0/45 & - & 4/27 \\
\bottomrule
\end{tabular}
\end{table}

\begin{sidewaystable}[htbp]
\centering
\caption{Structure learning experiments absolute log-likelihood error summary across different node structures, densities and sample sizes in structure learning experiments. Methods compared are DPM-SPBN with CAVI and GPCAVI inference methods, standalone DPM with CAVI and GPCAVI and SPBN with NR and UCV.}
\label{tab:abs_err_summary}
\tiny
\renewcommand{\arraystretch}{0.9}
\setlength{\tabcolsep}{3pt}
\begin{tabular}{@{}c p{1.8cm} l r@{.}l r@{.}l r@{.}l r@{.}l r@{.}l r@{.}l@{}}
\toprule
\textbf{Nodes} & \textbf{Density} & \textbf{\makecell{Sample \\ Size}} & \multicolumn{2}{c}{\textbf{\makecell{SPBN-CAVI \\ IQI}}} & \multicolumn{2}{c}{\textbf{\makecell{SPBN-GPCAVI \\ IQI}}} & \multicolumn{2}{c}{\textbf{\makecell{DPM-CAVI \\ IQI}}} & \multicolumn{2}{c}{\textbf{\makecell{DPM-GPCAVI \\ IQI}}} & \multicolumn{2}{c}{\textbf{\makecell{NR \\ IQI}}} & \multicolumn{2}{c}{\textbf{\makecell{UCV \\ IQI}}} \\
\midrule
\multirow{15}{*}{5} & \multirow{5}{1.8cm}{Smooth} & 200 & $3542$ & $61~[3489.32, 3822.96]$ & $\mathbf{3352}$ & $\mathbf{63}~[3221.55, 4194.58]$ & $6163$ & $79~[5812.99, 6936.79]$ & $6101$ & $40~[5861.03, 6591.19]$ & $6223$ & $77~[5821.42, 6940.49]$ & $9411$ & $32~[7658.25, 11396.03]$ \\
~ & ~ & 2000 & $\mathbf{989}$ & $\mathbf{43}~[930.75, 1140.64]$ & $1059$ & $55~[1009.85, 1143.88]$ & $1559$ & $82~[1505.76, 1658.67]$ & $1572$ & $77~[1485.10, 1631.78]$ & $3074$ & $67~[2907.57, 3096.99]$ & $2467$ & $37~[2217.27, 2550.98]$ \\
~ & ~ & 10000 & $443$ & $03~[424.78, 464.37]$ & $\mathbf{377}$ & $\mathbf{85}~[346.04, 386.45]$ & $560$ & $58~[530.06, 591.25]$ & $556$ & $46~[520.24, 563.34]$ & $1911$ & $02~[1814.85, 1923.92]$ & $1014$ & $75~[864.45, 1120.75]$ \\
~ & ~ & 50000 & $269$ & $21~[238.14, 286.06]$ & $\mathbf{199}$ & $\mathbf{39}~[188.22, 220.89]$ & $315$ & $18~[305.21, 320.16]$ & $259$ & $67~[252.88, 264.25]$ & $1127$ & $44~[1111.32, 1137.12]$ & \multicolumn{2}{c}{-} \\
~ & ~ & 100000 & $219$ & $16~[203.98, 235.41]$ & $\mathbf{173}$ & $\mathbf{52}~[167.41, 186.94]$ & $289$ & $03~[278.35, 298.28]$ & $244$ & $49~[228.24, 250.83]$ & $888$ & $74~[867.96, 904.21]$ & \multicolumn{2}{c}{-} \\
\cmidrule(l){2-15}
~ & \multirow{5}{1.8cm}{Medium} & 200 & $7054$ & $86~[6881.30, 7136.24]$ & $6993$ & $62~[6834.09, 7084.41]$ & $1$ & $17~[1.14, 1.22] \times 10^{4}$ & $1$ & $25~[1.17, 1.28] \times 10^{4}$ & $1$ & $03~[1.01, 1.05] \times 10^{4}$ & $\mathbf{3387}$ & $\mathbf{40}~[3224.21, 3561.86]$ \\
~ & ~ & 2000 & $\mathbf{1048}$ & $\mathbf{81}~[979.27, 1277.09]$ & $1225$ & $70~[1108.24, 1328.98]$ & $2693$ & $01~[2541.65, 2789.40]$ & $3326$ & $63~[2630.77, 3939.62]$ & $5484$ & $16~[5450.06, 5554.51]$ & $1482$ & $70~[1411.20, 1699.00]$ \\
~ & ~ & 10000 & $\mathbf{433}$ & $\mathbf{26}~[360.47, 476.54]$ & $490$ & $48~[432.40, 562.48]$ & $908$ & $77~[879.74, 994.62]$ & $1275$ & $03~[1020.24, 1789.09]$ & $3397$ & $33~[3368.25, 3442.81]$ & $656$ & $63~[643.91, 743.09]$ \\
~ & ~ & 50000 & $\mathbf{235}$ & $\mathbf{84}~[221.50, 248.90]$ & $257$ & $98~[215.33, 295.70]$ & $620$ & $52~[594.36, 676.06]$ & $1095$ & $26~[884.75, 1200.42]$ & $1904$ & $91~[1870.26, 1933.94]$ & \multicolumn{2}{c}{-} \\
~ & ~ & 100000 & $\mathbf{209}$ & $\mathbf{24}~[196.51, 216.52]$ & $220$ & $34~[200.28, 247.97]$ & $606$ & $83~[574.63, 626.47]$ & $1210$ & $60~[1138.74, 1276.07]$ & $1414$ & $08~[1365.12, 1439.65]$ & \multicolumn{2}{c}{-} \\
\cmidrule(l){2-15}
~ & \multirow{5}{1.8cm}{Rough} & 200 & $\mathbf{1}$ & $\mathbf{89}~[1.85, 1.94] \times 10^{4}$ & $1$ & $92~[1.84, 1.99] \times 10^{4}$ & $2$ & $29~[2.24, 2.37] \times 10^{4}$ & $2$ & $30~[2.26, 2.37] \times 10^{4}$ & $2$ & $26~[2.22, 2.34] \times 10^{4}$ & $2$ & $26~[2.25, 2.40] \times 10^{4}$ \\
~ & ~ & 2000 & $\mathbf{7095}$ & $\mathbf{64}~[6905.84, 7166.77]$ & $7301$ & $40~[7013.50, 7568.82]$ & $1$ & $05~[1.03, 1.07] \times 10^{4}$ & $1$ & $06~[1.04, 1.07] \times 10^{4}$ & $1$ & $75~[1.74, 1.76] \times 10^{4}$ & $8186$ & $92~[7147.64, 9331.70]$ \\
~ & ~ & 10000 & $\mathbf{3255}$ & $\mathbf{11}~[3180.77, 3345.69]$ & $3511$ & $99~[3386.41, 3655.57]$ & $6418$ & $48~[6303.76, 6520.14]$ & $7139$ & $97~[6444.20, 7573.78]$ & $1$ & $40~[1.39, 1.41] \times 10^{4}$ & $4457$ & $26~[4119.90, 5003.02]$ \\
~ & ~ & 50000 & $\mathbf{2062}$ & $\mathbf{69}~[1795.58, 2240.20]$ & $2557$ & $20~[2317.80, 2690.94]$ & $4843$ & $45~[4810.87, 5135.14]$ & $5355$ & $52~[5257.34, 5540.03]$ & $1$ & $09~[1.09, 1.10] \times 10^{4}$ & \multicolumn{2}{c}{-} \\
~ & ~ & 100000 & $\mathbf{1719}$ & $\mathbf{01}~[1537.62, 1914.13]$ & $1841$ & $43~[1714.65, 2015.47]$ & $4662$ & $15~[4547.28, 4851.14]$ & $4938$ & $35~[4845.12, 5022.84]$ & $9813$ & $84~[9768.88, 9898.00]$ & \multicolumn{2}{c}{-} \\
\midrule
\multirow{15}{*}{10} & \multirow{5}{1.8cm}{Smooth} & 200 & $\mathbf{4}$ & $\mathbf{76}~[4.63, 5.39] \times 10^{4}$ & $4$ & $86~[4.65, 5.45] \times 10^{4}$ & $6$ & $90~[6.66, 7.20] \times 10^{4}$ & $6$ & $93~[6.79, 7.22] \times 10^{4}$ & $1$ & $81~[1.36, 3.46] \times 10^{5}$ & $2$ & $23~[1.45, 3.45] \times 10^{5}$ \\
~ & ~ & 2000 & $\mathbf{2}$ & $\mathbf{57}~[2.38, 2.78] \times 10^{4}$ & $2$ & $64~[2.56, 3.06] \times 10^{4}$ & $4$ & $32~[4.08, 4.61] \times 10^{4}$ & $4$ & $32~[4.08, 4.61] \times 10^{4}$ & $1$ & $09~[0.83, 2.68] \times 10^{5}$ & $1$ & $24~[0.99, 4.86] \times 10^{5}$ \\
~ & ~ & 10000 & $\mathbf{1}$ & $\mathbf{72}~[1.60, 1.92] \times 10^{4}$ & $1$ & $82~[1.74, 2.03] \times 10^{4}$ & $3$ & $71~[3.32, 4.07] \times 10^{4}$ & $3$ & $70~[3.32, 4.06] \times 10^{4}$ & $1$ & $01~[0.98, 1.07] \times 10^{5}$ & $9$ & $44~[8.52, 10.38] \times 10^{4}$ \\
~ & ~ & 50000 & $1$ & $43~[1.37, 1.71] \times 10^{4}$ & $\mathbf{1}$ & $\mathbf{42}~[1.36, 1.61] \times 10^{4}$ & $3$ & $30~[3.14, 3.44] \times 10^{4}$ & $3$ & $26~[3.11, 3.40] \times 10^{4}$ & $6$ & $19~[5.70, 7.62] \times 10^{4}$ & \multicolumn{2}{c}{-} \\
~ & ~ & 100000 & $1$ & $32~[1.23, 1.44] \times 10^{4}$ & $\mathbf{1}$ & $\mathbf{24}~[1.22, 1.35] \times 10^{4}$ & $3$ & $03~[2.95, 3.05] \times 10^{4}$ & $3$ & $03~[2.94, 3.05] \times 10^{4}$ & $7$ & $24~[6.17, 7.66] \times 10^{4}$ & \multicolumn{2}{c}{-} \\
\cmidrule(l){2-15}
~ & \multirow{5}{1.8cm}{Medium} & 200 & $1$ & $32~[1.26, 1.35] \times 10^{4}$ & $1$ & $32~[1.30, 1.36] \times 10^{4}$ & $3$ & $41~[3.31, 3.46] \times 10^{4}$ & $3$ & $41~[3.31, 3.46] \times 10^{4}$ & $1$ & $74~[1.69, 1.81] \times 10^{4}$ & $\mathbf{1}$ & $\mathbf{07}~[1.03, 1.41] \times 10^{4}$ \\
~ & ~ & 2000 & $\mathbf{4136}$ & $\mathbf{80}~[3982.29, 4273.63]$ & $4654$ & $61~[4537.12, 4699.08]$ & $1$ & $37~[1.33, 1.42] \times 10^{4}$ & $1$ & $38~[1.34, 1.46] \times 10^{4}$ & $9632$ & $79~[9571.81, 9690.84]$ & $4970$ & $49~[4595.45, 5977.61]$ \\
~ & ~ & 10000 & $\mathbf{2276}$ & $\mathbf{16}~[2189.61, 2358.26]$ & $2421$ & $92~[2358.70, 2468.93]$ & $9509$ & $05~[9344.62, 9829.59]$ & $1$ & $04~[1.01, 1.10] \times 10^{4}$ & $6037$ & $64~[5976.46, 6212.55]$ & $2896$ & $78~[2808.14, 3005.70]$ \\
~ & ~ & 50000 & $\mathbf{1700}$ & $\mathbf{18}~[1662.51, 1764.97]$ & $1751$ & $55~[1649.47, 1786.14]$ & $8400$ & $47~[8140.49, 8692.28]$ & $8490$ & $16~[8022.05, 9770.59]$ & $3621$ & $03~[3594.92, 3649.58]$ & \multicolumn{2}{c}{-} \\
~ & ~ & 100000 & $1682$ & $36~[1620.55, 1773.70]$ & $\mathbf{1670}$ & $\mathbf{22}~[1626.16, 1712.85]$ & $8314$ & $63~[8272.12, 8423.61]$ & $8581$ & $54~[8244.05, 8856.83]$ & $2896$ & $77~[2868.68, 2946.29]$ & \multicolumn{2}{c}{-} \\
\cmidrule(l){2-15}
~ & \multirow{5}{1.8cm}{Rough} & 200 & $\mathbf{9}$ & $\mathbf{76}~[9.02, 9.95] \times 10^{4}$ & $9$ & $78~[9.16, 10.16] \times 10^{4}$ & $1$ & $16~[1.13, 1.18] \times 10^{5}$ & $1$ & $16~[1.13, 1.18] \times 10^{5}$ & $1$ & $84~[1.58, 3.35] \times 10^{5}$ & $1$ & $87~[1.73, 2.54] \times 10^{5}$ \\
~ & ~ & 2000 & $\mathbf{7}$ & $\mathbf{29}~[7.09, 7.44] \times 10^{4}$ & $7$ & $41~[7.17, 7.54] \times 10^{4}$ & $10$ & $00~[9.89, 10.12] \times 10^{4}$ & $9$ & $95~[9.83, 10.09] \times 10^{4}$ & $1$ & $29~[1.23, 1.48] \times 10^{5}$ & $1$ & $57~[1.52, 1.64] \times 10^{5}$ \\
~ & ~ & 10000 & $\mathbf{6}$ & $\mathbf{04}~[5.95, 6.08] \times 10^{4}$ & $6$ & $09~[5.99, 6.12] \times 10^{4}$ & $9$ & $01~[8.82, 9.15] \times 10^{4}$ & $8$ & $99~[8.82, 9.15] \times 10^{4}$ & $1$ & $10~[1.09, 1.13] \times 10^{5}$ & $1$ & $44~[1.35, 1.54] \times 10^{5}$ \\
~ & ~ & 50000 & $5$ & $07~[5.02, 5.16] \times 10^{4}$ & $\mathbf{4}$ & $\mathbf{96}~[4.92, 5.00] \times 10^{4}$ & $8$ & $01~[7.79, 8.20] \times 10^{4}$ & $7$ & $95~[7.86, 8.06] \times 10^{4}$ & $9$ & $98~[9.96, 10.12] \times 10^{4}$ & \multicolumn{2}{c}{-} \\
~ & ~ & 100000 & $4$ & $63~[4.61, 4.67] \times 10^{4}$ & $\mathbf{4}$ & $\mathbf{60}~[4.57, 4.62] \times 10^{4}$ & $7$ & $77~[7.70, 7.89] \times 10^{4}$ & $7$ & $78~[7.72, 7.84] \times 10^{4}$ & $9$ & $64~[9.62, 9.68] \times 10^{4}$ & \multicolumn{2}{c}{-} \\
\midrule
\multirow{15}{*}{15} & \multirow{5}{1.8cm}{Smooth} & 200 & $1$ & $81~[1.75, 1.87] \times 10^{4}$ & $1$ & $89~[1.85, 1.99] \times 10^{4}$ & $6$ & $67~[6.52, 6.80] \times 10^{4}$ & $6$ & $66~[6.52, 6.81] \times 10^{4}$ & $2$ & $69~[2.67, 2.76] \times 10^{4}$ & $\mathbf{1}$ & $\mathbf{44}~[1.39, 1.85] \times 10^{4}$ \\
~ & ~ & 2000 & $6997$ & $72~[6692.65, 7259.37]$ & $7300$ & $29~[7036.60, 7492.82]$ & $1$ & $97~[1.91, 2.00] \times 10^{4}$ & $2$ & $00~[1.95, 2.05] \times 10^{4}$ & $1$ & $68~[1.67, 1.70] \times 10^{4}$ & $\mathbf{5910}$ & $\mathbf{04}~[5781.78, 6466.65]$ \\
~ & ~ & 10000 & $\mathbf{2932}$ & $\mathbf{49}~[2837.43, 3017.17]$ & $3189$ & $21~[3143.43, 3324.55]$ & $1$ & $06~[1.03, 1.11] \times 10^{4}$ & $1$ & $07~[1.06, 1.11] \times 10^{4}$ & $1$ & $27~[1.26, 1.28] \times 10^{4}$ & $3694$ & $44~[3510.51, 4292.74]$ \\
~ & ~ & 50000 & $\mathbf{1629}$ & $\mathbf{48}~[1576.26, 1753.59]$ & $1891$ & $99~[1801.89, 1970.44]$ & $8474$ & $11~[8364.13, 8548.26]$ & $8680$ & $59~[8577.23, 8768.77]$ & $9372$ & $15~[9318.36, 9428.45]$ & \multicolumn{2}{c}{-} \\
~ & ~ & 100000 & $\mathbf{1445}$ & $\mathbf{99}~[1360.02, 1461.06]$ & $1623$ & $56~[1544.42, 1701.35]$ & $8329$ & $79~[8182.18, 8473.34]$ & $8408$ & $27~[8218.09, 8495.85]$ & $8318$ & $39~[8287.55, 8409.88]$ & \multicolumn{2}{c}{-} \\
\cmidrule(l){2-15}
~ & \multirow{5}{1.8cm}{Medium} & 200 & $\mathbf{1}$ & $\mathbf{74}~[1.73, 1.78] \times 10^{5}$ & $1$ & $76~[1.73, 1.81] \times 10^{5}$ & $2$ & $42~[2.39, 2.45] \times 10^{5}$ & $2$ & $42~[2.39, 2.45] \times 10^{5}$ & $2$ & $31~[1.80, 3.17] \times 10^{6}$ & $7$ & $18~[6.19, 19.84] \times 10^{5}$ \\
~ & ~ & 2000 & $\mathbf{1}$ & $\mathbf{36}~[1.33, 1.43] \times 10^{5}$ & $1$ & $40~[1.36, 1.43] \times 10^{5}$ & $2$ & $13~[2.10, 2.16] \times 10^{5}$ & $2$ & $13~[2.10, 2.16] \times 10^{5}$ & $2$ & $89~[2.58, 10.65] \times 10^{5}$ & $2$ & $83~[2.46, 3.73] \times 10^{5}$ \\
~ & ~ & 10000 & $\mathbf{1}$ & $\mathbf{26}~[1.22, 1.32] \times 10^{5}$ & $1$ & $26~[1.22, 1.34] \times 10^{5}$ & $2$ & $12~[2.10, 2.13] \times 10^{5}$ & $2$ & $12~[2.10, 2.13] \times 10^{5}$ & $2$ & $47~[2.21, 2.61] \times 10^{5}$ & $2$ & $44~[2.30, 3.24] \times 10^{5}$ \\
~ & ~ & 50000 & $1$ & $11~[1.10, 1.13] \times 10^{5}$ & $\mathbf{1}$ & $\mathbf{10}~[1.09, 1.15] \times 10^{5}$ & $2$ & $06~[2.02, 2.09] \times 10^{5}$ & $2$ & $06~[2.02, 2.07] \times 10^{5}$ & $2$ & $03~[1.83, 2.15] \times 10^{5}$ & \multicolumn{2}{c}{-} \\
~ & ~ & 100000 & $1$ & $10~[1.09, 1.12] \times 10^{5}$ & $\mathbf{1}$ & $\mathbf{09}~[1.07, 1.11] \times 10^{5}$ & $2$ & $01~[1.99, 2.02] \times 10^{5}$ & $2$ & $00~[2.00, 2.02] \times 10^{5}$ & $2$ & $03~[2.02, 2.10] \times 10^{5}$ & \multicolumn{2}{c}{-} \\
\cmidrule(l){2-15}
~ & \multirow{5}{1.8cm}{Rough} & 200 & $\mathbf{2}$ & $\mathbf{63}~[2.61, 2.64] \times 10^{5}$ & $2$ & $63~[2.63, 2.64] \times 10^{5}$ & $3$ & $11~[3.10, 3.16] \times 10^{5}$ & $3$ & $11~[3.10, 3.16] \times 10^{5}$ & $3$ & $33~[3.24, 3.42] \times 10^{5}$ & $3$ & $38~[3.34, 3.45] \times 10^{5}$ \\
~ & ~ & 2000 & $\mathbf{2}$ & $\mathbf{21}~[2.21, 2.23] \times 10^{5}$ & $2$ & $23~[2.22, 2.24] \times 10^{5}$ & $2$ & $89~[2.88, 2.90] \times 10^{5}$ & $2$ & $89~[2.88, 2.90] \times 10^{5}$ & $2$ & $81~[2.79, 2.83] \times 10^{5}$ & $3$ & $07~[3.01, 3.16] \times 10^{5}$ \\
~ & ~ & 10000 & $2$ & $05~[2.04, 2.05] \times 10^{5}$ & $\mathbf{2}$ & $\mathbf{05}~[2.04, 2.05] \times 10^{5}$ & $2$ & $72~[2.71, 2.73] \times 10^{5}$ & $2$ & $72~[2.71, 2.73] \times 10^{5}$ & $2$ & $64~[2.63, 2.66] \times 10^{5}$ & $2$ & $86~[2.82, 3.05] \times 10^{5}$ \\
~ & ~ & 50000 & $\mathbf{1}$ & $\mathbf{90}~[1.90, 1.91] \times 10^{5}$ & $1$ & $91~[1.89, 1.91] \times 10^{5}$ & $2$ & $59~[2.58, 2.60] \times 10^{5}$ & $2$ & $58~[2.57, 2.60] \times 10^{5}$ & $2$ & $51~[2.50, 2.52] \times 10^{5}$ & \multicolumn{2}{c}{-} \\
~ & ~ & 100000 & $\mathbf{1}$ & $\mathbf{85}~[1.85, 1.86] \times 10^{5}$ & $1$ & $86~[1.85, 1.87] \times 10^{5}$ & $2$ & $54~[2.54, 2.54] \times 10^{5}$ & $2$ & $54~[2.54, 2.54] \times 10^{5}$ & $2$ & $46~[2.46, 2.46] \times 10^{5}$ & \multicolumn{2}{c}{-} \\
\midrule
\multicolumn{3}{c}{\textbf{Wins / Total}} & \multicolumn{2}{c}{29/45} & \multicolumn{2}{c}{12/45} & \multicolumn{2}{c}{0/45} & \multicolumn{2}{c}{0/45} & \multicolumn{2}{c}{0/45} & \multicolumn{2}{c}{4/27} \\
\bottomrule
\end{tabular}
\end{sidewaystable}

\subsection{Real data experiments}
In this section, we evaluate the performance of the proposed DPM-SPBN on real-world datasets, for which no ground truth is known. For this purpose, we used a selection of datasets from the UCI Machine Learning Repository, chosen to provide variability in the number of variables and sample sizes (see Table \ref{uci_tables}).

\begin{table}[htbp]
\centering
\caption{Datasets from the Machine Learning UCI repository.}
\label{uci_tables}
\renewcommand{\arraystretch}{1.2}
\setlength{\tabcolsep}{8pt}
\begin{tabular}{@{}l l l@{}}
    \toprule
    \textbf{Dataset} & $N$ & \textbf{\# Variables} \\
    \midrule
    Glass                           & $214$    & $9$  \\
    Ionosphere                      & $351$    & $31$ \\
    QSAR Aquatic                    & $546$    & $9$  \\
    Breast Cancer                   & $683$    & $9$  \\
    QSAR Fish Toxicity              & $908$    & $7$  \\
    Yeast                           & $1484$  & $8$  \\
    Wine Quality-White              & $4898$  & $12$ \\
    Waveform                        & $5000$  & $40$ \\
    Travel Ratings                  & $5546$  & $24$ \\
    Magic Gamma                     & $19020$ & $10$ \\
    Gas Turbine CO and NOx Emission & $29349$ & $11$ \\
    MetroPT-3                       & $50000$ & $7$  \\
    \bottomrule
\end{tabular}
\end{table}

Given the results of the previous sections, where the SPBN framework outperforms its competitors, we focus on this framework and compare the standard bandwidth selectors NR and UCV with the proposed DPM-SPBN using CAVI and GPCAVI for inference. Since the underlying network structure is unknown for the real-world datasets, we use HC with cross-validation score for structure learning for all methods. As in Section~\ref{sec:structure_experiments}, UCV results are not reported for the MetroPT and Gas Emission datasets due to high computational cost.

\begin{figure}[htbp]
    \centering
    \includegraphics[width=0.8\textwidth]{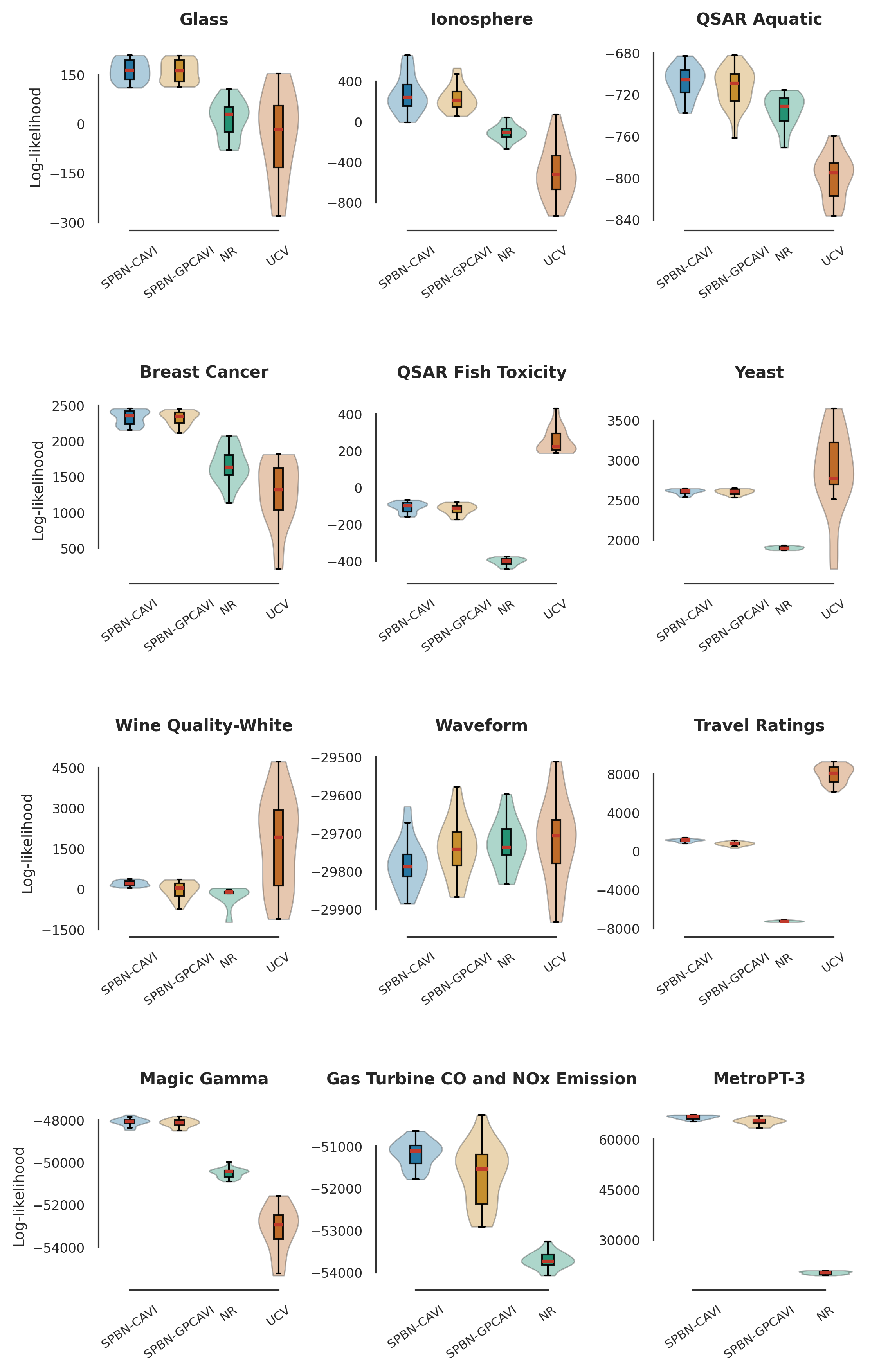}
    \caption{Performance comparison on real-world datasets between the proposed DPM-SPBN with inference methods CAVI and GPCAVI and state-of-the-art SPBNs using NR and UCV bandwidth selection. Distributions show test-set log-likelihoods obtained from $10$-fold cross-validation.}
    \label{fig:real_data_results}
\end{figure}

Since the underlying data generating distribution is unknown for real-world datasets, we cannot compute the absolute LL error. Instead, we use LL as the performance criterion because it reflects how well a model fits the available data, which makes it suitable for model comparison in this setting. However, LL can favor models that fit the observed data very closely while providing suboptimal density approximations due to overfitting which has to be considered. Nevertheless, LL is widely used for BN learning and evaluation, where mechanisms are incorporated to mitigate overfitting. For each dataset, we perform $10$-fold cross-validation and report the resulting LL distributions in Figure~\ref{fig:real_data_results}. The experimental setup, including the DPM hyperparameters, is the same as in the previous sections. For structure learning, we consider a grid of hyperparameters for the HC score, as described in Section~\ref{sec:structure_experiments}. Hyperparameters are selected according to the same criterion, namely the best median LL.

The results demonstrate the general superiority in performance of the proposed DPM-SPBN across most datasets, regardless of the inference method used. SPBN-CAVI and SPBN-GPCAVI perform similarly, although GPCAVI yields slightly worse results in a few isolated cases. NR is consistently outperformed in all scenarios except for the Waveform dataset, where all four methods achieve a similar median LL. In particular, the DPM-SPBN methods exhibit remarkably tight and stable distributions across folds compared to the baselines. However, UCV outperforms the proposed methods in four specific settings (QSAR Fish Toxicity, Yeast, Wine Quality-White and Travel Ratings). This shows that UCV remains a highly competitive baseline that cannot be discarded in scenarios with low-to-medium sample sizes and variables, despite its substantially higher variance across folds. These results are broadly consistent with previous experiments.

\section{Conclusions and future research}
\label{sec:5}

In this paper, we introduced a novel CPD modeling approach for continuous variables based on Bayesian non-parametric DPMs and integrated it into the SPBN framework. The proposed DPM-SPBN combines the flexibility of SPBNs for continuous-variable modeling with the adaptive density representation of DPMs. We considered three variational inference strategies, CAVI, GPCAVI, and Memoized, providing different trade-offs between computational cost and density-estimation performance.

The experimental results demonstrate that DPM-SPBN provides an effective approach for JPD modeling of continuous variables and clearly outperforms the state-of-the-art alternatives considered in this study. Among the proposed inference strategies, CAVI provides the most consistent overall performance, achieving strong results in both CPD estimation and structure learning while maintaining a favorable computational cost. GPCAVI provides broadly similar performance and, therefore, represents a viable alternative with comparable computational characteristics. Based on the results obtained in this study, CAVI and GPCAVI provide the most practical choices for the proposed DPM-SPBN framework, particularly when both parameter estimation and structure learning are considered.

Memoized provides an important complementary result. Its adaptive truncation can improve the density-estimation performance for larger datasets and more complex distributions, demonstrating the potential benefits of allowing the mixture complexity to adapt to the data. However, its substantially higher computational cost becomes a major limitation when CPDs must be repeatedly fitted during HC structure learning. Consequently, Memoized was beneficial primarily when the network structure was fixed and could not be incorporated into the full HC experiments considered in this work. This limitation motivates further research into structure-learning methods that can be combined with Memoized CPD learning as the final parameter estimation method.

The comparison with standalone DPMs further demonstrates the benefit of the Bayesian network factorization, with DPM-SPBN consistently achieving better density estimation under the experimental conditions considered. Together with the results obtained on real-world datasets, these findings support DPM-SPBN as an effective extension of SPBNs for continuous-variable JPD modeling, combining flexible conditional density estimation with the computational and representational advantages of BN factorization.

Several directions remain open for future research. First, the proposed framework could be extended to additional data settings, including online, mixed discrete and continuous, and time-sliced data. This would require inference procedures tailored to these settings, such as the approach of \cite{yang2019memorized} for online data. Second, a more comprehensive comparison of alternative DPM inference methods could provide further insight into the trade-offs between computational efficiency, uncertainty modeling, interpretability, and density-estimation performance. For example, hard-clustering methods could provide faster training, but their effect on density estimation performance would need to be investigated. Finally, future work should investigate alternative structure learning procedures, including approaches that are agnostic or specifically designed for the underlying CPD model. Furthermore, studying the relationship between CPD estimation performance, structure learning methods, and sstructural recovery could identify methods that better exploit the properties of DPM-based CPDs, including memoization, and thus reduce the computational cost of structure learning.

\section*{Declaration of generative AI and AI-assisted technologies in the writing process.}
During the preparation of this work, the authors used ChatGPT to improve the coherence and clarity of the text. After using this service, the authors reviewed and edited the content as needed and assume full responsibility for the content of the published article.

\section*{Research data statement}
The datasets used in this work include publicly available UCI datasets, which can be accessed through the  \href{https://archive.ics.uci.edu/}{UCI Machine Learning Repository}. Additionally, synthetic datasets developed by the authors, as well as the modified PyBNesian package and the associated experiments code, are available in the first author's GitHub repository at https://github.com/Victor-Alejandre/DPM-SPBN.git. All data and code used in this study are publicly accessible for reproducibility.

\section*{Acknowledgments}
The authors gratefully acknowledge the Universidad Politécnica de Madrid (www.upm.es) for providing computing resources on Magerit Supercomputer.

\section*{Funding sources}
This work was partially supported by the Ministry of Science, Innovation and Universities, Spain under Project AEI/MICIU/10.13039/501100011033/PID2025-170500NB-100 and Project PLEC2023-010252/MIG-20232016. Also, by the Autonomous Region of Madrid under Project ELLIS Unit Madrid and TEC-2024/COM-89.

% Usar el estilo numérico por orden de aparición
\bibliographystyle{unsrt}
\bibliography{references}

@article{ferguson1973bayesian,
  title={A {B}ayesian analysis of some nonparametric problems},
  author={Ferguson, Thomas S},
  journal={The Annals of Statistics},
  pages={209--230},
  year={1973},
  publisher={JSTOR}
}

@article{antoniak1974mixtures,
  title={Mixtures of {D}irichlet processes with applications to {B}ayesian nonparametric problems},
  author={Antoniak, Charles E},
  journal={The Annals of Statistics},
  pages={1152--1174},
  year={1974},
  publisher={JSTOR}
}

@article{sethuraman1994constructive,
  title={A constructive definition of {D}irichlet priors},
  author={Sethuraman, Jayaram},
  journal={Statistica Sinica},
  pages={639--650},
  year={1994},
  publisher={JSTOR}
}

@article{blackwell1973ferguson,
  title={Ferguson distributions via {P}{\'o}lya urn schemes},
  author={Blackwell, David and MacQueen, James B},
  journal={The Annals of Statistics},
  volume={1},
  number={2},
  pages={353--355},
  year={1973},
  publisher={Institute of Mathematical Statistics}
}

@article{jordan_blei,
author = {David M. Blei and Michael I. Jordan},
title = {{Variational inference for {D}irichlet process mixtures}},
volume = {1},
journal = {{B}ayesian {A}nalysis},
number = {1},
publisher = {International Society for {B}ayesian {A}nalysis},
pages = {121 -- 143},
year = {2006},
doi = {10.1214/06-BA104}
}

@article{hoffman,
  author  = {Matthew D. Hoffman and David M. Blei and Chong Wang and John Paisley},
  title   = {Stochastic variational inference},
  journal = {Journal of {M}achine {L}earning {R}esearch},
  year    = {2013},
  volume  = {14},
  number  = {40},
  pages   = {1303--1347}
}

@article{wang2011fast,
  title={Fast {B}ayesian inference in {D}irichlet process mixture models},
  author={Wang, Lianming and Dunson, David B},
  journal={Journal of Computational and Graphical Statistics},
  volume={20},
  number={1},
  pages={196--216},
  year={2011},
  publisher={Taylor \& Francis}
}

@article{Zhang02102014,
author = {Xiaole Zhang and David J. Nott and Christopher Yau and Ajay Jasra},
title = {A Sequential Algorithm for Fast Fitting of {D}irichlet Process Mixture Models},
journal = {Journal of {C}omputational and {G}raphical {S}tatistics},
volume = {23},
number = {4},
pages = {1143--1162},
year = {2014},
publisher = {Taylor \& Francis}
}

@book{koller2009probabilistic,
  title={Probabilistic {G}raphical {M}odels: {P}rinciples and {T}echniques},
  author={Koller, Daphne and Friedman, Nir},
  year={2009},
  publisher={The MIT Press}
}

@article{gaussiannetworks,
author = {Shachter, Ross D. and Kenley, C. Robert},
title = {Gaussian Influence Diagrams},
year = {1989},
issue_date = {May 1989},
publisher = {INFORMS},
volume = {35},
number = {5},
journal = {Management Science},
pages = {527–550}
}

@article{masmoudi2019new,
  title={A new class of continuous {B}ayesian networks},
  author={Masmoudi, Khalil and Masmoudi, Afif},
  journal={International Journal of Approximate Reasoning},
  volume={109},
  pages={125--138},
  year={2019},
  publisher={Elsevier}
}

@inproceedings{moral2001mixtures,
  title={Mixtures of truncated exponentials in hybrid {B}ayesian networks},
  author={Moral, Seraf{\'\i}n and Rum{\'\i}, Rafael and Salmer{\'o}n, Antonio},
  booktitle={Proceedings of the 6th European Conference, Symbolic and Quantitative Approaches to Reasoning with Uncertainty},
  pages={156--167},
  year={2001},
  organization={Springer}
}

@article{shenoy2011inference,
  title={Inference in hybrid {B}ayesian networks using mixtures of polynomials},
  author={Shenoy, Prakash P and West, James C},
  journal={International Journal of Approximate Reasoning},
  volume={52},
  number={5},
  pages={641--657},
  year={2011},
  publisher={Elsevier}
}

@article{langseth2012mixtures,
  title={Mixtures of truncated basis functions},
  author={Langseth, Helge and Nielsen, Thomas D and Rum{\'{i}}, Rafael and Salmer{\'o}n, Antonio},
  journal={International Journal of Approximate Reasoning},
  volume={53},
  number={2},
  pages={212--227},
  year={2012},
  publisher={Elsevier}
}

@article{cooper1992bayesian,
  title={A {B}ayesian method for the induction of probabilistic networks from data},
  author={Cooper, Gregory F and Herskovits, Edward},
  journal={Machine Learning},
  volume={9},
  pages={309--347},
  year={1992},
  publisher={Springer}
}

@inproceedings{Friedman2000GaussianPN,
author = {Friedman, Nir and Nachman, Iftach},
title = {Gaussian Process Networks},
year = {2000},
isbn = {1558607099},
publisher = {Morgan Kaufmann Publishers Inc.},
booktitle = {Proceedings of the 16th Conference on Uncertainty in Artificial Intelligence},
pages = {211–219},
numpages = {9}
}

@article{ickstadt2010nonparametric,
  title={Nonparametric {B}ayesian networks},
  author={Ickstadt, Katja and Bornkamp, Bj{\"o}rn and Grzegorczyk, Marco and Wieczorek, Jakob and Sheriff, M Rahuman and Grecco, Hernan E and Zamir, Eli},
  journal={Bayesian Statistics},
  volume={9},
  pages={283--316},
  year={2010},
  publisher={Oxford Univ. Press}
}

@article{hofmann1995discovering,
  title={Discovering structure in continuous variables using {B}ayesian networks},
  author={Hofmann, Reimar and Tresp, Volker},
  journal={Advances in Neural Information Processing Systems},
  volume={8},
  year={1995},
  pages={500--506}
}

@article{atienza2022semiparametric,
  title={Semiparametric {B}ayesian networks},
  author={Atienza, David and Bielza, Concha and Larra{\~n}aga, Pedro},
  journal={Information Sciences},
  volume={584},
  pages={564--582},
  year={2022},
  publisher={Elsevier}
}

@article{atienza2022pybnesian,
  title={Py{BN}esian: An extensible {P}ython package for {B}ayesian networks},
  author={Atienza, David and Bielza, Concha and Larra{\~n}aga, Pedro},
  journal={Neurocomputing},
  volume={504},
  pages={204--209},
  year={2022},
  publisher={Elsevier}
}

@article{kurowicka2005distribution,
  title={Distribution-free continuous {B}ayesian belief nets},
  author={Kurowicka, Dorota and Cooke, RM},
  journal={Modern Statistical and Mathematical Methods in Reliability},
  volume={10},
  pages={309--323},
  year={2005},
  publisher={World Scientific Publishing Hackensack, NJ}
}

@article{hanea2015non,
  title={Non-parametric {B}ayesian networks: Improving theory and reviewing applications},
  author={Hanea, Anca and Napoles, Oswaldo Morales and Ababei, Dan},
  journal={Reliability Engineering \& System Safety},
  volume={144},
  pages={265--284},
  year={2015},
  publisher={Elsevier}
}

@article{bauer2016pair,
  title={Pair-copula {B}ayesian networks},
  author={Bauer, Alexander and Czado, Claudia},
  journal={Journal of Computational and Graphical Statistics},
  volume={25},
  number={4},
  pages={1248--1271},
  year={2016},
  publisher={Taylor \& Francis}
}

@article{elidan2010copula,
  title={Copula {B}ayesian networks},
  author={Elidan, Gal},
  journal={Advances in Neural Information Processing Systems},
  volume={23},
  year={2010}
}

@article{lasserre2021constraint,
  title={Constraint-based learning for non-parametric continuous {B}ayesian networks},
  author={Lasserre, Marvin and Lebrun, R{\'e}gis and Wuillemin, Pierre-Henri},
  journal={Annals of Mathematics and Artificial Intelligence},
  volume={89},
  number={10},
  pages={1035--1052},
  year={2021},
  publisher={Springer}
}

@article{Boukabour17122021,
author = {Seloua Boukabour and Afif Masmoudi},
title = {Semiparametric {B}ayesian networks for continuous data},
journal = {Communications in Statistics - Theory and Methods},
volume = {50},
number = {24},
pages = {5974--5996},
year = {2021},
publisher = {Taylor \& Francis}

}

@article{atienza2022hybrid,
  title={Hybrid semiparametric {B}ayesian networks},
  author={Atienza, David and Larra{\~n}aga, Pedro and Bielza, Concha},
  journal={Test},
  volume={31},
  number={2},
  pages={299--327},
  year={2022},
  publisher={Springer}
}

@article{colombo2014order,
  title={Order-independent constraint-based causal structure learning},
  author={Colombo, Diego and Maathuis, Marloes H},
  journal={Journal of Machine Learning Research},
  volume={15},
  number={1},
  pages={3741--3782},
  year={2014},
  publisher={JMLR. org}
}

@article{alejandre2026bandwidth,
title = {Bandwidth selectors on semiparametric {B}ayesian networks},
journal = {Information Sciences},
volume = {740},
pages = {123198},
year = {2026},
issn = {0020-0255},
doi = {https://doi.org/10.1016/j.ins.2026.123198},
author={Alejandre, Victor and Bielza, Concha and Larra{\~n}aga}
}

@article{raykov,
author = {Yordan P. Raykov and Alexis Boukouvalas and Max A. Little},
title = {{Simple approximate MAP inference for Dirichlet processes mixtures}},
volume = {10},
journal = {Electronic Journal of Statistics},
number = {2},
publisher = {Institute of Mathematical Statistics and Bernoulli Society},
pages = {3548 -- 3578},
year = {2016},
doi = {10.1214/16-EJS1196}
}

@article{neal2000markov,
  title={Markov chain sampling methods for {D}irichlet process mixture models},
  author={Neal, Radford M},
  journal={Journal of Computational and Graphical Statistics},
  volume={9},
  number={2},
  pages={249--265},
  year={2000},
  publisher={Taylor \& Francis}
}

@article{ishwaran2001gibbs,
  title={Gibbs sampling methods for stick-breaking priors},
  author={Ishwaran, Hemant and James, Lancelot F},
  journal={Journal of the American Statistical Association},
  volume={96},
  number={453},
  pages={161--173},
  year={2001},
  publisher={Taylor \& Francis}
}

@article{kalli2011slice,
  title={Slice sampling mixture models},
  author={Kalli, Maria and Griffin, Jim E and Walker, Stephen G},
  journal={Statistics and Computing},
  volume={21},
  number={1},
  pages={93--105},
  year={2011},
  publisher={Springer}
}

@inproceedings{ulker2010sequential,
  title={Sequential {M}onte {C}arlo samplers for {D}irichlet process mixtures},
  author={Ulker, Yener and G{\"u}nsel, Bilge and Cemgil, Taylan},
  booktitle={Proceedings of the Thirteenth International Conference on Artificial Intelligence and Statistics},
  pages={876--883},
  year={2010},
  organization={JMLR Workshop and Conference Proceedings}
}

@article{hastie2015sampling,
  title={Sampling from {D}irichlet process mixture models with unknown concentration parameter: {M}ixing issues in large data implementations},
  author={Hastie, David I and Liverani, Silvia and Richardson, Sylvia},
  journal={Statistics and Computing},
  volume={25},
  number={5},
  pages={1023--1037},
  year={2015},
  publisher={Springer}
}

@article{jain2004split,
  title={A split-merge {M}arkov chain {M}onte {C}arlo procedure for the Dirichlet process mixture model},
  author={Jain, Sonia and Neal, Radford M},
  journal={Journal of Computational and Graphical Statistics},
  volume={13},
  number={1},
  pages={158--182},
  year={2004},
  publisher={Taylor \& Francis}
}

@article{maceachern1998estimating,
  title={Estimating mixture of {D}irichlet process models},
  author={MacEachern, Steven N and M{\"u}ller, Peter},
  journal={Journal of Computational and Graphical Statistics},
  volume={7},
  number={2},
  pages={223--238},
  year={1998},
  publisher={Taylor \& Francis}
}

@inproceedings{kurihara2007collapsed,
  title={Collapsed variational {D}irichlet Process Mixture Models.},
  author={Kurihara, Kenichi and Welling, Max and Teh, Yee Whye},
  booktitle={Ijcai},
  volume={7},
  pages={2796--2801},
  year={2007}
}

@article{hughes2013memoized,
  title={Memoized online variational inference for {D}irichlet process mixture models},
  author={Hughes, Michael C and Sudderth, Erik},
  journal={Advances in Neural Information Processing Systems},
  volume={26},
  year={2013}
}

@article{lin2013online,
  title={Online learning of nonparametric mixture models via sequential variational approximation},
  author={Lin, Dahua},
  journal={Advances in Neural Information Processing Systems},
  volume={26},
  year={2013}
}

@article{wang2012truncation,
  title={Truncation-free online variational inference for {B}ayesian nonparametric models},
  author={Wang, Chong and Blei, David},
  journal={Advances in Neural Information Processing Systems},
  volume={25},
  year={2012}
}

@article{campbell2015streaming,
  title={Streaming, distributed variational inference for {B}ayesian nonparametrics},
  author={Campbell, Trevor and Straub, Julian and Fisher III, John W and How, Jonathan P},
  journal={Advances in Neural Information Processing Systems},
  volume={28},
  year={2015}
}

@article{yang2019memorized,
  title={Memorized variational continual learning for {D}irichlet process mixtures},
  author={Yang, Yang and Chen, Bo and Liu, Hongwei},
  journal={IEEE Access},
  volume={7},
  pages={150851--150862},
  year={2019},
  publisher={IEEE}
}

@inproceedings{huynh2016streaming,
  title={Streaming variational inference for {D}irichlet process mixtures},
  author={Huynh, Viet and Phung, Dinh and Venkatesh, Svetha},
  booktitle={Asian Conference on Machine Learning},
  pages={237--252},
  year={2016},
  organization={PMLR}
}

@article{cgbnlaurizten,
author = {S. L. Lauritzen and N. Wermuth},
title = {{Graphical models for associations between variables, some of which are qualitative and some quantitative}},
volume = {17},
journal = {The Annals of Statistics},
number = {1},
publisher = {Institute of Mathematical Statistics},
pages = {31 -- 57},
year = {1989},
doi = {10.1214/aos/1176347003}
}

@article{friedman2000using,
  title={Using {B}ayesian networks to analyze expression data},
  author={Friedman, N. and Linial, M. and Nachman, I. and Pe'er, D.},
  journal={Journal of Computational Biology},
  volume={7},
  number={3-4},
  pages={601--620},
  year={2000},
  doi={10.1089/106652700750050961}
}

@article{kyrimi2021comprehensive,
  title={A comprehensive scoping review of {B}ayesian networks in healthcare: Past, present and future},
  author={Kyrimi, E. and McLachlan, S. and Dube, K. and Neves, M. R. and Fahmi, A. and Fenton, N.},
  journal={Artificial Intelligence in Medicine},
  volume={117},
  pages={102108},
  year={2021},
  doi={10.1016/j.artmed.2021.102108}
}

@article{bielza2014bayesian,
  title={Bayesian networks in neuroscience: {A} survey},
  author={Bielza, C. and Larra{\~n}aga, P.},
  journal={Frontiers in Computational Neuroscience},
  volume={8},
  pages={131},
  year={2014},
  doi={10.3389/fncom.2014.00131}
}

@article{WEBER2012671,
title = {Overview on {B}ayesian networks applications for dependability, risk analysis and maintenance areas},
journal = {Engineering Applications of Artificial Intelligence},
volume = {25},
number = {4},
pages = {671-682},
year = {2012},
issn = {0952-1976},
doi = {https://doi.org/10.1016/j.engappai.2010.06.002},
author = {P. Weber and G. Medina-Oliva and C. Simon and B. Iung}
}

@article{Wand1992,
author = {M. P. Wand},
title = {Error analysis for general multtvariate kernel estimators},
journal = {Journal of Nonparametric Statistics},
volume = {2},
number = {1},
pages = {1--15},
year = {1992},
publisher = {Taylor \& Francis},
doi = {10.1080/10485259208832538}
}

@article{quintana2022dependent,
author = {Fernando A. Quintana and Peter M{\"u}ller and Alejandro Jara and Steven N. MacEachern},
title = {{The dependent {D}irichlet process and related models}},
volume = {37},
journal = {Statistical Science},
number = {1},
publisher = {Institute of Mathematical Statistics},
pages = {24 -- 41},
year = {2022}
}

@article{benavoli2016should,
  title={Should we really use post-hoc tests based on mean-ranks?},
  author={Benavoli, Alessio and Corani, Giorgio and Mangili, Francesca},
  journal={Journal of Machine Learning Research},
  volume={17},
  number={1},
  pages={152--161},
  year={2016},
  publisher={JMLR. org}
}

@article{garcia2008extension,
  author  = {Salvador Garc{{\'i}}a and Francisco Herrera},
  title   = {An Extension on ``{S}tatistical Comparisons of Classifiers over Multiple Data Sets'' for all Pairwise Comparisons},
  journal = {Journal of Machine Learning Research},
  year    = {2008},
  volume  = {9},
  number  = {89},
  pages   = {2677--2694}
}

@article{bonnini2024review,
  title={Review about the permutation approach in hypothesis testing},
  author={Bonnini, Stefano and Assegie, Getnet Melak and Trzcinska, Kamila},
  journal={Mathematics},
  volume={12},
  number={17},
  pages={2617},
  year={2024},
  publisher={MDPI}
}

@article{tsamardinos2006max,
  title={The max-min hill-climbing {B}ayesian network structure learning algorithm},
  author={Tsamardinos, Ioannis and Brown, Laura E and Aliferis, Constantin F},
  journal={Machine Learning},
  volume={65},
  pages={31--78},
  year={2006},
  publisher={Springer}
}

@book{chacon2018multivariate,
  title={Multivariate {K}ernel {S}moothing and its {A}pplications},
  author={Chac{\'o}n, Jos{\'e} E and Duong, Tarn},
  year={2018},
  publisher={Chapman and Hall/CRC}
}

@article{kitson2023survey,
  title={A survey of {B}ayesian Network structure learning},
  author={Kitson, Neville Kenneth and Constantinou, Anthony C and Guo, Zhigao and Liu, Yang and Chobtham, Kiattikun},
  journal={Artificial Intelligence Review},
  volume={56},
  number={8},
  pages={8721--8814},
  year={2023},
  publisher={Springer}
}

@article{Maceachern1994,
author = {Steven N. Maceachern},
title = {Estimating normal means with a conjugate style {D}irichlet process prior},
journal = {Communications in Statistics, Simulation and Computation},
volume = {23},
number = {3},
pages = {727--741},
year = {1994},
publisher = {Taylor \& Francis},
doi = {10.1080/03610919408813196}

}

\end{document}